\pdfoutput=1

\documentclass[11pt]{article}

\usepackage[final]{acl}

\usepackage{times}
\usepackage{latexsym}

\usepackage[T1]{fontenc}

\usepackage{graphicx}
\usepackage{subcaption}

\usepackage[utf8]{inputenc}

\usepackage{microtype}

\usepackage{inconsolata}

\usepackage{graphicx}
\usepackage{xcolor}
\usepackage{colortbl}

\author{Ho Yin `Sam' Ng\textsuperscript{1}~~~~
Ting-Yao Hsu\textsuperscript{1}~~~~
Aashish Anantha Ramakrishnan\textsuperscript{1}\\
\textbf{Branislav Kveton\textsuperscript{2}~~~~
Nedim Lipka\textsuperscript{2}~~~~
Franck Dernoncourt\textsuperscript{2}~~~~
Dongwon Lee\textsuperscript{1}}\\
\textbf{Tong Yu\textsuperscript{2}~~~~
Sungchul Kim\textsuperscript{2}~~~~
Ryan A. Rossi\textsuperscript{2}~~~~
Ting-Hao `Kenneth' Huang\textsuperscript{1}}\\
  \textsuperscript{1}The Pennsylvania State University~~~~
  \textsuperscript{2}Adobe Research\\
\textsuperscript{1}\texttt{\{sam.ng,txh357,aashish,dongwon,txh710\}@psu.edu}~~~\\
\textsuperscript{2}\texttt{\{kveton,lipka,dernonco,tyu,sukim,ryrossi\}@adobe.com}\\
}

\usepackage{xspace}
\usepackage{comment}
\usepackage{xcolor} 
\usepackage{booktabs}
\usepackage{siunitx}    
\usepackage{tablefootnote}  
\usepackage{threeparttable} 

\usepackage{amsmath}
\usepackage{cleveref}
\usepackage{multirow}
\usepackage{fvextra} 

\usepackage[utf8]{inputenc} 
\usepackage{CJKutf8}
\usepackage{siunitx}
\usepackage{listings}
\usepackage{tabularx}
\usepackage{enumitem}

\usepackage{soul}
\usepackage{hyperref}

\PassOptionsToPackage{table,xcdraw}{xcolor}
\usepackage{colortbl}

\definecolor{lightred}{RGB}{255,200,200}
\definecolor{lightblue}{RGB}{200,200,255}
\definecolor{lightgreen}{RGB}{200,255,200}
\definecolor{lightyellow}{RGB}{255,255,200}
\definecolor{eclipseStrings}{RGB}{42,0.0,255}
\definecolor{eclipseKeywords}{RGB}{127,0,85}
\colorlet{numb}{magenta!60!black}

\newcommand{\redbold}[1]{\textcolor{red}}

\title{\dataset: Personalized Figure Caption Generation\\With Multimodal Figure Profiles}

\newcommand{\eg}{{\it e.g.}\xspace}
\newcommand{\ie}{{\it i.e.}\xspace}

\newcommand{\dataset}{\mbox{\textsc{LaMP-Cap}}\xspace}

\begin{document}
\maketitle

\begin{abstract}
Figure captions are crucial for helping readers understand and remember a figure's key message. 
Many models have been developed to generate these captions, helping authors compose better quality captions more easily.
Yet, authors almost always need to revise generic AI-generated captions to match their writing style and the domain's style, highlighting the need for personalization.
Despite language models' personalization (LaMP) advances, these technologies often focus on text-only settings and rarely address scenarios where both inputs and profiles are multimodal.
This paper introduces \textbf{\dataset},\footnote{Data:~\href{https://github.com/Crowd-AI-Lab/lamp-cap}{https://github.com/Crowd-AI-Lab/lamp-cap}} a dataset for personalized figure caption generation with multimodal figure profiles.
For each target figure, \dataset provides not only the needed inputs, such as figure images, but also up to three other figures from the same document---each with its image, caption, and figure-mentioning paragraphs---as a \textit{profile} to characterize the context.
Experiments with four LLMs show that using profile information consistently helps generate captions closer to the original author-written ones.
Ablation studies reveal that images in the profile are more helpful than figure-mentioning paragraphs, highlighting the advantage of using multimodal profiles over text-only ones.

\end{abstract}

\section{Introduction\label{sec:intro}}
\begin{figure}[t]
    \centering
    \includegraphics[width=.99\columnwidth]{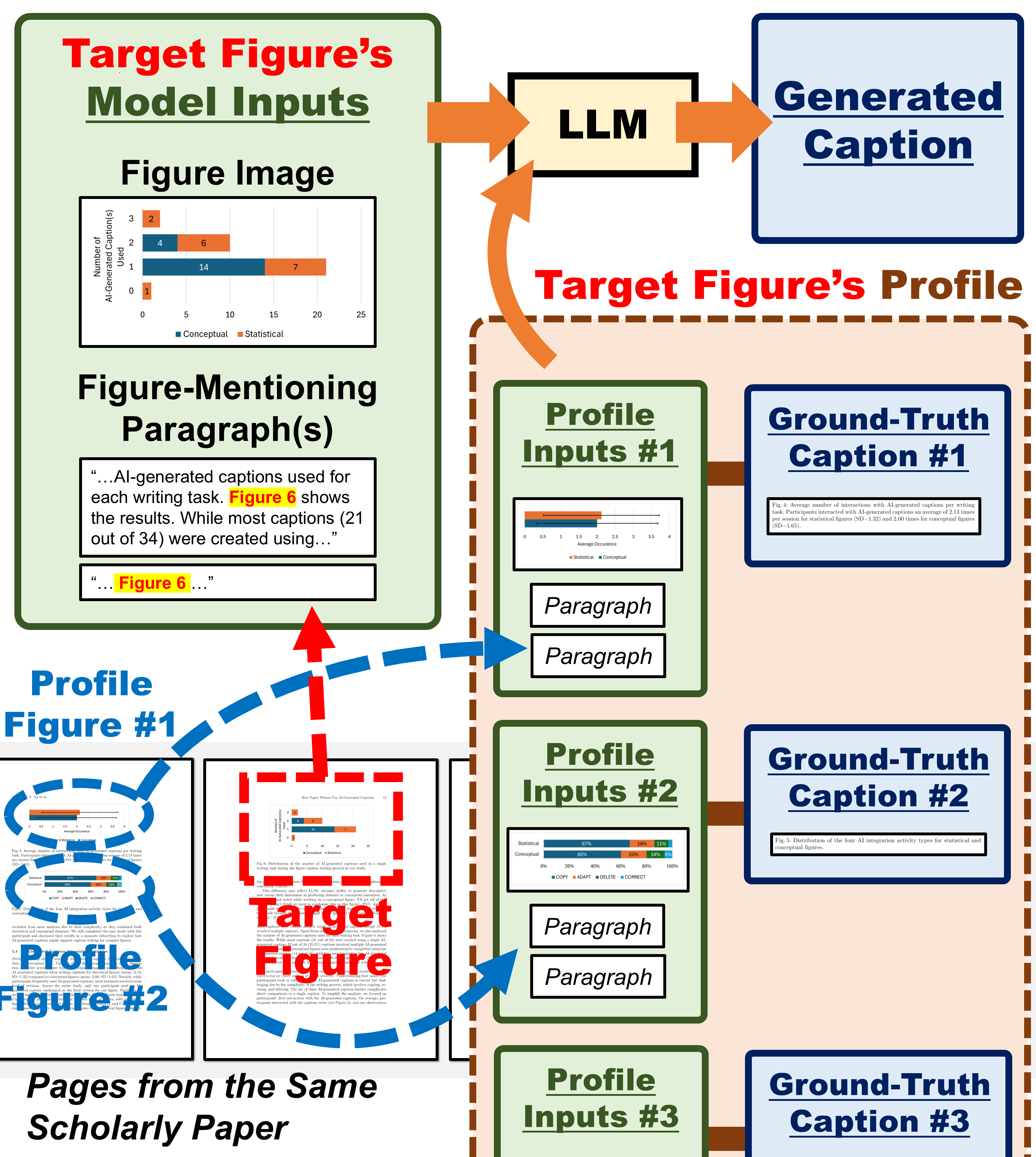}
    \vspace{-1.5pc}
    \caption{Overview of \dataset. For each target figure, the dataset provides multimodal \textit{inputs}---the figure image and figure-mentioning paragraphs---and a multimodal \textit{profile} of up to three other figures (\ie, profile figures) from the same paper, each with its image, caption, and related paragraphs. The model generates a caption for the target figure using the inputs and profile.}
    \vspace{-1pc}
    \label{fig:overview}
\end{figure}

Figures like bar charts or line charts are widely used by scientists, companies, and governments to communicate key insights~\cite{kim2021towards,farahani2023automatic}.
Captions---text placed next to these figures---are known to be crucial for helping readers understand and remember the figure's message~\cite{tang2023vistext,kantharaj2022chart,meng-etal-2024-chartassistant}.
Many models have been developed to generate high-quality captions to help authors compose captions more easily~\cite{hsu-etal-2021-scicap-generating,huang-etal-2023-summaries,liu-etal-2023-matcha,masry2023unichart}.
For example, the \textsc{SciCap} Challenges in 2023 and 2024 invited global teams to generate captions for scientific figures in arXiv papers~\cite{hsu2025large,kim2025multi}.
Systems like \textsc{SciCapenter} also emerged to assist authors by providing AI-generated captions~\cite{10.1145/3613905.3650738}.
Despite these advances, studies show that authors almost always need to revise generic AI-generated captions to match their style and the domain's style, with one participant noting, ``I need to revise the facade because this is \textit{not the right way} to present (the concept)''~\cite{ng2025understanding,ngunderstanding}.
This highlights the need for personalized caption generation.

Meanwhile, the rise of large language models (LLMs) has recently fueled interest in personalized text generation~\cite{zhang2024personalization,wozniak2024personalized}. 
Benchmarks like \textsc{LaMP}~\cite{salemi-etal-2024-lamp} (\ul{\textsc{La}}nguage \ul{\textsc{M}}odels \ul{\textsc{P}}ersonalization) and \textsc{LongLaMP}~\cite{kumar2024longlamp} were created to study how LLMs can tailor text for specific contexts.
However, most of these explorations focused on text-only settings, where both the input (used for generation) and profile (used for personalization) were text-based.
How these text-only approaches apply to multimodal scenarios---such as figure caption generation---remains unclear.

This paper introduces \textbf{\dataset}, a dataset for \textbf{personalized figure caption generation with multimodal figure profiles} (\S\ref{sec:dataset}).
\dataset includes 110,828 target figures---scientific figures for which models aim to generate captions for---each from a distinct arXiv paper.
For each target figure, \dataset provides the needed inputs (\textit{source})---figure images and figure-mentioning paragraphs (\eg, ``Figure 3 shows...'')---along with up to three other figures from the same paper, each with its image, caption, and figure-mentioning paragraphs, as a \textit{profile} to capture context.
Models are then tasked with generating captions for the target figure using its image and figure-mentioning paragraphs (multimodal \textit{source}), given a figure profile of source-caption pairs from the same paper (multimodal \textit{profile} for personalization).
We used \dataset to test caption generation with four LLMs and found that profile information consistently improved the similarity of generated captions to ground-truth captions (\S\ref{sec:results}).
Ablation studies revealed that captions are the most critical profile element, followed by images, with figure-mentioning paragraphs being the least important (\S\ref{sec:ablation}).
Our work provides a new benchmark for personalized text generation and demonstrates the effectiveness of using multimodal profiles beyond text-only approaches.

\section{Related Work\label{sec:realted-work}}
\paragraph{Figure Caption Generation.}
Figure caption generation requires models to understand both the visual content and the broader context~\cite{kantharaj-etal-2022-chart,wang2024charxiv,hu2024mplug,obeid-hoque-2020-chart}.
Early approaches, like \textsc{FigCAP} and the initial version of \textsc{SciCap}, relied solely on figure images as input~\cite{chen2020figure,hsu-etal-2021-scicap-generating}.
Researchers soon realized this was insufficient and began incorporating additional context, such as figure-mentioning paragraphs and even the document's title or abstract~\cite{huang-etal-2023-summaries,yang2024scicap+,stokes2022striking}.
Despite this progress, prior work often overlooked personalization. 
Although studies noted that users often need captions tailored to their style or domain~\cite{hsu2025large,huang-etal-2023-summaries},
none of these approaches explicitly provided source-target pairs 
that capture the specific generation context needed for models to learn personalized styles. 
A few studies have explored creative personalization of image captions~\cite{Shuster2019-mi, Anantha-Ramakrishnan2025-je}, but these approaches relied on explicit style inputs, making them dependent on user-provided style descriptions.


\paragraph{Personalized LLMs.}
Personalization of LLMs has gained attention~\cite{zhang2024personalization}, primarily in two directions: 
{\em (i)} personalized text generation (tailoring generated text for specific contexts) and 
{\em (ii)} downstream task personalization (enhancing targeted applications like recommendation systems).
We focus on the first direction, defining the personalization target as a \textit{group} of users---all co-authors of a paper---rather than individuals. 
Prior work, such as the LaMP-5 task~\cite{salemi-etal-2024-lamp} on Personalized Scholarly Title Generation, has also treated a paper's author group as a single entity for personalization.
Most prior work in this space has been centered on text-only settings (\S\ref{sec:intro}).
For example, \textsc{LaMP} included tasks such as news headline generation and email subject creation---relying exclusively on text-based inputs and profiles~\cite{salemi-etal-2024-lamp}.
How these approaches extend to multimodal scenarios remains an open question. 



\section{\dataset Dataset\label{sec:dataset}}
We constructed \dataset by curating the \textsc{SciCap} Challenge Dataset~\cite{hsu2025large}.
We first selected all papers containing at least two figures. 
From each paper, we then randomly designated one figure as the \textbf{target figure} (the one needing a caption) and used the remaining figures from that paper (up to a maximum of three, since the \textsc{SciCap} Challenge allowed at most four figures per paper) as the \textbf{profile} to provide personalization context.

Following the \textsc{SciCap} Challenge Dataset's split (\ie, 80/10/10 train/val/test), \dataset includes 110,828 target figures: 
86,197 for training, 
12,361 for validation, and 
12,270 for testing. 
Among these, 54,680 (49.3\%) had one profile figure, 26,193 (23.6\%) had two, and 30,027 (27.1\%) had three, totaling 197,075 profile figures. 
Papers with only one figure were excluded. 
See \autoref{app:dataset-details} for details.

\section{Experimental Results\label{sec:results}}


\paragraph{Experiment Setups.}\label{para:Experiment-Setups}
We evaluated four LLMs
on personalized caption generation using \dataset:\footnote{Qwen-2.5-VL-7B-Instruct was excluded from our main analysis due to significantly higher failure rate (2.2\%, 269/12,259) compared to other models. See~\autoref{app:output-cleaning}.}
{\em (i)} GPT-4o~\cite{hurst2024gpt}, 
{\em (ii)} Llama 4 Scout~\cite{meta2025llama4}, 
{\em (iii)} Gemini 2.5 Flash Preview~\cite{gemini25flash}, and 
{\em (iv)} GPT-4.1 Mini~\cite{openai2024gpt41mini}.
The first three are larger models, while the last one is smaller.
We used OpenAI's API for GPT-4o and OpenRouter (openrouter.ai) for the others. 
We focused on LLMs because large-scale human evaluations from the \textsc{SciCap} Challenge showed a clear performance gap between model classes~\cite{hsu2025large}: only LLMs like GPT-4V consistently generate captions matching or exceeding those by human authors, while smaller or specialized models such as PEGASUS~\cite{zhang2020pegasus} and UniChart~\cite{masry2023unichart} perform poorly.


Building on prior work showing that more profile information improves performance~\cite{tan2024democratizing}, we tested four caption generation settings with varying amounts and sources of profile input:
\textbf{(1) No Profile:} The model generated captions using only the target figure's image and figure-mentioning paragraphs.
\textbf{(2) One Profile:} The model used the same source as in (1) but additionally used \textit{one} randomly selected profile figure from the \textit{same} paper
for personalization.
\textbf{(3) All Profile:} The model used the same source as in (1) but additionally used \textit{all} profile figures from the \textit{same} paper for personalization.
\textbf{(4) Other Profile:} 
The model used the same source as in (1) but additionally used
one or three randomly selected profile figure(s) from random \textit{other} papers. 
The setup (4) tests whether performance gains come from paper-specific context 
or generic in-domain examples.
See~\autoref{app:prompts} for the full prompt.

We cleaned the output by removing unnecessary reasoning steps or explanations. 
We also removed cases (56 out of 12,259) where models failed to generate valid output. 
See \autoref{app:output-cleaning} and 
\autoref{app:text-evaluation} for details.

\begin{table}[t]
\centering
\footnotesize
\renewcommand{\arraystretch}{1.215}
\setlength{\tabcolsep}{2.0pt}
\definecolor{vlightgreen}{rgb}{0.95, 1.0, 0.95}
\definecolor{lightgreen}{rgb}{0.85, 0.95, 0.85}
\definecolor{darkgreen}{rgb}{0.7, 0.9, 0.7}

\definecolor{lightred}{rgb}{1.0, 0.9, 0.9}
\definecolor{darkred}{rgb}{0.9, 0.7, 0.7}






\begin{tabular}{@{}lccccccccc@{}}
\toprule
 &
\multicolumn{2}{c}{\textbf{Profile}} &
  \multicolumn{4}{c}{\textbf{BLEU}} &
  \multicolumn{3}{c}{\textbf{ROUGE}} \\ \cmidrule(l){2-10} 
\multirow{-2}{*}{\textbf{LLM}} &
  \textbf{\begin{tabular}[c]{@{}c@{}}Same\\ Paper\end{tabular}} &
  \textbf{\begin{tabular}[c]{@{}c@{}}\#\\ $\le 3$\end{tabular}} &
  \textbf{B-1} &
  \textbf{B-2} &
  \textbf{B-3} &
  \textbf{B-4} &
  \textbf{R-1} &
  \textbf{R-2} &
  \textbf{R-L} \\ \midrule
 &
  N/A &
  0 &
  .219 &
  .133 &
  .091 &
  .063 &
  .321 &
  .127 &
  .248 \\ \cmidrule(l){2-10} 
 &
   &
  \cellcolor{darkred}1 &
  \cellcolor{darkred}.189 &
  \cellcolor{darkred}.108 &
  \cellcolor{darkred}.070 &
  \cellcolor{darkred}.046 &
  \cellcolor{darkred}.289 &
  \cellcolor{darkred}.101 &
  \cellcolor{darkred}.218 \\
 &
  \multirow{-2}{*}{N} &
  \cellcolor{lightred}3 &
  \cellcolor{lightred}.206 &
  \cellcolor{lightred}.118 &
  \cellcolor{lightred}.076 &
  \cellcolor{lightred}.051 &
  \cellcolor{lightred}.300 &
  \cellcolor{lightred}.107 &
  \cellcolor{lightred}.223 \\ \cmidrule(l){2-10} 
 &
   &
  \cellcolor{lightgreen}1 &
  \cellcolor{lightgreen}.279 &
  \cellcolor{lightgreen}.186 &
  \cellcolor{lightgreen}.137 &
  \cellcolor{lightgreen}.103 &
  \cellcolor{lightgreen}.384 &
  \cellcolor{lightgreen}.178 &
  \cellcolor{lightgreen}.313 \\
\multirow{-5}{*}{\begin{tabular}[c]{@{}l@{}}GPT-\\ 4o\end{tabular}} &
  \multirow{-2}{*}{Y} &
    \cellcolor{darkgreen} All &
  \cellcolor{darkgreen}.292 &
  \cellcolor{darkgreen}.200 &
  \cellcolor{darkgreen}.150 &
  \cellcolor{darkgreen}.115 &
  \cellcolor{darkgreen}.397 &
  \cellcolor{darkgreen}.194 &
  \cellcolor{darkgreen}.328 \\ \midrule
 &
  N/A &
  0 &
  .254 &
  .178 &
  .138 &
  .112 &
  .357 &
  .182 &
  .293 \\ \cmidrule(l){2-10} 
 &
   &
  \cellcolor{darkred}1 &
  \cellcolor{darkred}.223 &
  \cellcolor{darkred}.167 &
  \cellcolor{darkred}.132 &
  \cellcolor{darkred}.107 &
  \cellcolor{darkred}.348 &
  \cellcolor{darkred}.180 &
  \cellcolor{darkred}.295 \\
 &
  \multirow{-2}{*}{N} &
  \cellcolor{vlightgreen}3 &
  \cellcolor{vlightgreen}.251 &
  \cellcolor{vlightgreen}.187 &
  \cellcolor{vlightgreen}.152 &
  \cellcolor{vlightgreen}.129 &
  \cellcolor{vlightgreen}.368 &
  \cellcolor{vlightgreen}.202 &
  \cellcolor{vlightgreen}.317 \\ \cmidrule(l){2-10} 
 &
   &
  \cellcolor{lightgreen}1 &
  \cellcolor{lightgreen}.372 &
  \cellcolor{lightgreen}.293 &
  \cellcolor{lightgreen}.246 &
  \cellcolor{lightgreen}.211 &
  \cellcolor{lightgreen}.481 &
  \cellcolor{lightgreen}.300 &
  \cellcolor{lightgreen}.423 \\
\multirow{-5}{*}{\begin{tabular}[c]{@{}l@{}}Llama-\\ 4 \\ Scout\end{tabular}} &
  \multirow{-2}{*}{Y} &
  \cellcolor{darkgreen}All &
  \cellcolor{darkgreen}.396 &
  \cellcolor{darkgreen}.318 &
  \cellcolor{darkgreen}.270 &
  \cellcolor{darkgreen}.235 &
  \cellcolor{darkgreen}.503 &
  \cellcolor{darkgreen}.324 &
  \cellcolor{darkgreen}.447 \\ \midrule
 &
  N/A &
  0 &
  .305 &
  .230 &
  .188 &
  .160 &
  .417 &
  .237 &
  .361 \\ \cmidrule(l){2-10} 
 &
   &
  \cellcolor{darkred}1 &
  \cellcolor{darkred}.268 &
  \cellcolor{darkred}.198 &
  \cellcolor{darkred}.159 &
  \cellcolor{darkred}.132 &
  \cellcolor{darkred}.388 &
  \cellcolor{darkred}.213 &
  \cellcolor{darkred}.335 \\
 &
  \multirow{-2}{*}{N} &
  \cellcolor{lightred}3 &
  \cellcolor{lightred}.279 &
  \cellcolor{lightred}.205 &
  \cellcolor{lightred}.164 &
  \cellcolor{lightred}.137 &
  \cellcolor{lightred}.398 &
  \cellcolor{lightred}.217 &
  \cellcolor{lightred}.343 \\ \cmidrule(l){2-10} 
 &
   &
  \cellcolor{lightgreen}1 &
  \cellcolor{lightgreen}.370 &
  \cellcolor{lightgreen}.291 &
  \cellcolor{lightgreen}.244 &
  \cellcolor{lightgreen}.209 &
  \cellcolor{lightgreen}.482 &
  \cellcolor{lightgreen}.301 &
  \cellcolor{lightgreen}.426 \\
\multirow{-5}{*}{\begin{tabular}[c]{@{}l@{}}Gemini-\\ 2.5\\ Flash \\ Preview\end{tabular}} &
  \multirow{-2}{*}{Y} &
  \cellcolor{darkgreen}All &
  \cellcolor{darkgreen}.395 &
  \cellcolor{darkgreen}.317 &
  \cellcolor{darkgreen}.270 &
  \cellcolor{darkgreen}.234 &
  \cellcolor{darkgreen}.504 &
  \cellcolor{darkgreen}.328 &
  \cellcolor{darkgreen}.449 \\ \midrule
 &
  N/A &
  0 &
  .209 &
  .124 &
  .081 &
  .054 &
  .305 &
  .117 &
  .225 \\ \cmidrule(l){2-10} 
 &
   &
  \cellcolor{vlightgreen}1 &
  \cellcolor{vlightgreen}.216 &
  \cellcolor{vlightgreen}.133 &
  \cellcolor{vlightgreen}.091 &
  \cellcolor{vlightgreen}.063 &
  \cellcolor{vlightgreen}.325 &
  \cellcolor{vlightgreen}.132 &
  \cellcolor{vlightgreen}.250 \\
 &
  \multirow{-2}{*}{N} &
  \cellcolor{vlightgreen}3 &
  \cellcolor{vlightgreen}.220 &
  \cellcolor{vlightgreen}.136 &
  \cellcolor{vlightgreen}.093 &
  \cellcolor{vlightgreen}.065 &
  \cellcolor{vlightgreen}.328 &
  \cellcolor{vlightgreen}.134 &
  \cellcolor{vlightgreen}.252 \\ \cmidrule(l){2-10} 
 &
   &
  \cellcolor{lightgreen}1 &
  \cellcolor{lightgreen}.286 &
  \cellcolor{lightgreen}.202 &
  \cellcolor{lightgreen}.155 &
  \cellcolor{lightgreen}.121 &
  \cellcolor{lightgreen}.398 &
  \cellcolor{lightgreen}.207 &
  \cellcolor{lightgreen}.326 \\
\multirow{-5}{*}{\begin{tabular}[c]{@{}l@{}}GPT-\\ 4.1 \\ Mini\end{tabular}} &
  \multirow{-2}{*}{Y} &
  \cellcolor{darkgreen}All &
  \cellcolor{darkgreen}.300 &
  \cellcolor{darkgreen}.218 &
  \cellcolor{darkgreen}.171 &
  \cellcolor{darkgreen}.137 &
  \cellcolor{darkgreen}.412 &
  \cellcolor{darkgreen}.225 &
  \cellcolor{darkgreen}.342 \\ \bottomrule
\end{tabular}

\vspace{-.5pc}
\caption{Performance of LLMs on caption generation across profile settings. The highest scores are achieved by using \textbf{all} available profile(s) from the \textbf{same} paper.}
\vspace{-1pc}
\label{tab:llm_comparison}
\end{table}



\paragraph{Using profile information (from the same paper) makes captions more similar to ground truth, especially with all profile figures.}\label{para:profile-result}
\autoref{tab:llm_comparison} shows the personalized caption generation results of four LLMs, evaluated using BLEU~\cite{papineni2002bleu} and ROUGE~\cite{lin2004rouge}.\footnote{We also explored BERTScore~\cite{zhang2019bertscore}, which correlated highly with BLEU and ROUGE. See \autoref{app:main-study-full}.}
We used reference-based metrics to measure how closely the generated captions matched the original author-written captions, following a standard evaluation approach for personalized text generation used in well-known work like LongLaMP~\cite{kumar2024longlamp}.
The results show that incorporating profile information consistently improves caption quality across all four models.
Additionally, using all profile figures provides better results than using just one.
See \autoref{app:main-study-full} for details.


\begin{figure*}[t]
    \centering
    \begin{subfigure}{0.495\textwidth}
        \renewcommand\thesubfigure{a-1}
        \includegraphics[width=\textwidth]{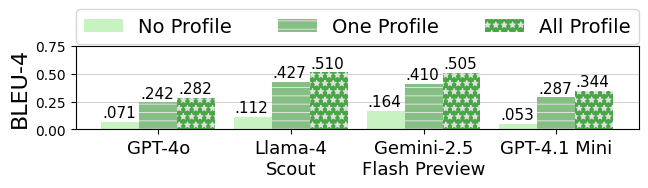}
        \vspace{-1.7pc}
        \caption{BLEU-4 (Context-Aligned Subset)}
        \label{fig:top_left}
    \end{subfigure}
    \hfill
    \begin{subfigure}{0.495\textwidth}
        \renewcommand\thesubfigure{b-1}
        \includegraphics[width=\textwidth]{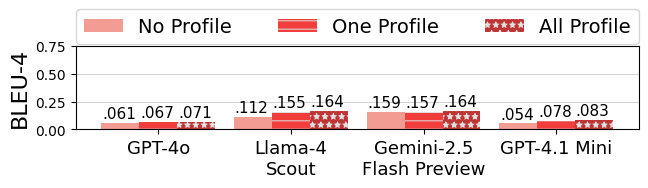}
        \vspace{-1.7pc}
        \caption{BLEU-4 (Context-Misaligned Subset)}
        \label{fig:top_right}
    \end{subfigure}
    
    \begin{subfigure}{0.495\textwidth}
         \renewcommand\thesubfigure{a-2}
        \includegraphics[width=\textwidth]{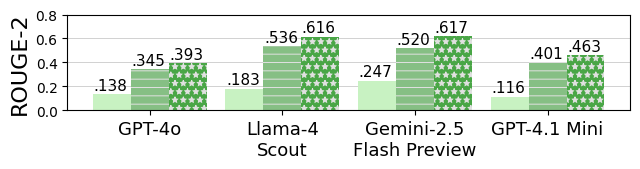}
        \vspace{-1.7pc}
        \caption{ROUGE-2 (Context-Misaligned Subset)}
        \label{fig:bottom_left}
    \end{subfigure}
    \hfill
    \begin{subfigure}{0.495\textwidth}
        \renewcommand\thesubfigure{b-2}
        \includegraphics[width=\textwidth]{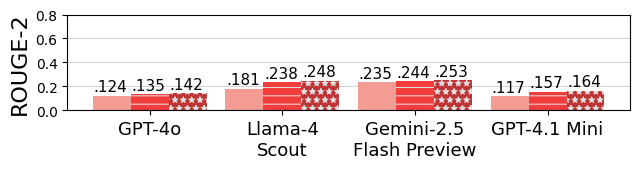}
        \vspace{-1.7pc}
        \caption{ROUGE-2 (Context-Misaligned Subset)}
        \label{fig:bottom_right}
    \end{subfigure}
    \vspace{-1.5pc}
    \caption{BLEU-4 and ROUGE-2 scores on \dataset's Context-Aligned and Context-Misaligned subsets, showing that personalization is most effective when profile caption are similar to the target (a-1, a-2).}
    \vspace{-1pc}
    \label{fig:all_figures}
\end{figure*}


\paragraph{Using profiles from other papers often lowered BLEU and ROUGE scores, though there were exceptions.} 
\autoref{tab:llm_comparison} suggests that performance gains primarily came from paper-specific context, rather than generic in-domain examples. 
Profiles from other papers generally hurt performance, but some models, such as GPT-4.1 Mini, showed slight improvements. 
Furthermore, using more \textit{other} profiles tended to reduce the performance drop or, occasionally, provide minor gains.

\begin{table}
\centering
\footnotesize
\renewcommand{\arraystretch}{1.2}
\setlength{\tabcolsep}{2pt}
\definecolor{lightgreen}{rgb}{0.9, 1.0, 0.9}
\definecolor{darkgreen}{rgb}{0.7, 0.9, 0.7}
\begin{tabular}{lcccccccc}
\toprule
\multirow{2}{*}{\textbf{LLM}} & \multirow{2}{*}{\textbf{Same}} & \multicolumn{4}{c}{\textbf{BLEU}} & \multicolumn{3}{c}{\textbf{ROUGE}} \\ 
\cmidrule(lr){3-6} \cmidrule(lr){7-9}
 & \textbf{Type} & \textbf{B-1} & \textbf{B-2} & \textbf{B-3} & \textbf{B-4} & \textbf{R-1} & \textbf{R-2} & \textbf{R-L} \\ 
\midrule
\multirow{2}{*}{GPT-4o} 
 & No & .233 & .142 & .097 & .069 & .340 & .134 & .270 \\ 
 & \cellcolor{darkgreen}Yes & \cellcolor{darkgreen}.302 & \cellcolor{darkgreen}.208 & \cellcolor{darkgreen}.157 & \cellcolor{darkgreen}.121 & \cellcolor{darkgreen}.406 & \cellcolor{darkgreen}.201 & \cellcolor{darkgreen}.336 \\ 
\midrule
\multirow{2}{*}{Llama-4 Scout}
 & No & .325 & .245 & .199 & .167 & .436 & .250 & .377 \\ 
 & \cellcolor{darkgreen}Yes & \cellcolor{darkgreen}.396 & \cellcolor{darkgreen}.317 & \cellcolor{darkgreen}.269 & \cellcolor{darkgreen}.233 & \cellcolor{darkgreen}.504 & \cellcolor{darkgreen}.325 & \cellcolor{darkgreen}.447 \\ 
\midrule
\multirow{2}{*}{\begin{tabular}[c]{@{}l@{}}Gemini-2.5\\Flash Preview\end{tabular}} 
 & No & .326 & .246 & .201 & .169 & .440 & .254 & .384 \\ 
 & \cellcolor{darkgreen}Yes & \cellcolor{darkgreen}.393 & \cellcolor{darkgreen}.314 & \cellcolor{darkgreen}.266 & \cellcolor{darkgreen}.229 & \cellcolor{darkgreen}.503 & \cellcolor{darkgreen}.325 & \cellcolor{darkgreen}.447 \\ 
\midrule
\multirow{2}{*}{GPT-4.1 Mini}  
 & No & .237 & .154 & .109 & .079 & .349 & .155 & .276 \\ 
 & \cellcolor{darkgreen}Yes & \cellcolor{darkgreen}.311 & \cellcolor{darkgreen}.227 & \cellcolor{darkgreen}.178 & \cellcolor{darkgreen}.142 & \cellcolor{darkgreen}.422 & \cellcolor{darkgreen}.234 & \cellcolor{darkgreen}.351 \\ 
\bottomrule
\end{tabular}
\vspace{-.5pc}
\caption{LLM performance on figures with one profile figure. Personalization is more effective when the single profile figure shares the same type as the target.}
\vspace{-1pc}
\label{tab:main-result-sameType}
\end{table}

\paragraph{When profile figures shared the same type as the target figure, personalization works better.}
To examine how figure type affects personalization, we analyzed cases with a single profile figure, splitting them into two groups: those where the profile and target figure types matched (n=8,083) and those where they did not (n=4,120). \autoref{tab:main-result-sameType} shows that matching the figure type resulted in captions that were significantly closer to the gold caption.


\paragraph{Personalization is more effective when profile captions are highly similar to the target caption.}\label{para:context-alignment}

To test if personalization is more effective when profiles are similar to the target, we split our test set into two groups. We calculated the similarity (using BERTScore and ROUGE-L) between each target caption and its available profile captions. The top 25\% of examples with the most similar profiles formed our Context-Aligned set (n=2,513); the remainder formed the Context-Misaligned set. The results in \autoref{fig:all_figures} confirm our hypothesis. Performance gains from using profiles were substantially larger for the Context-Aligned group~(Figures~\ref{fig:top_left} and~\ref{fig:bottom_left}), while the impact was noticeably smaller for the Context-Misaligned set~((Figures~\ref{fig:top_right} and~\ref{fig:bottom_right}). 
(See \autoref{app:easy-hard}.)

\begin{table}[t]
\centering
\footnotesize
\renewcommand{\arraystretch}{1.2}
\setlength{\tabcolsep}{2pt}
\definecolor{lightgreen}{rgb}{0.9, 1.0, 0.9}
\definecolor{darkgreen}{rgb}{0.7, 0.9, 0.7}


\begin{tabular}{@{}clllllll@{}}
\toprule
\multirow{2}{*}{\begin{tabular}[c]{@{}c@{}}Profile\\ \#\end{tabular}} & \multicolumn{4}{c}{BLEU}                                                                              & \multicolumn{3}{c}{ROUGE}                                                   \\ \cmidrule(l){2-8} 
                                                                      & \multicolumn{1}{c}{B-1} & \multicolumn{1}{c}{B-2} & \multicolumn{1}{c}{B-3} & \multicolumn{1}{c}{B-4} & \multicolumn{1}{c}{R-1} & \multicolumn{1}{c}{R-2} & \multicolumn{1}{c}{R-L} \\ \midrule
0                                                                     & .212                    & .124                    & .082                     & .055                    & .302                    & .113                    & .223                    \\
1                                                                     & .289                    & .198                    & .145                    & .109                    & .390                    & .181                    & .312                    \\
2                                                                     & .319                    & .215                    & .159                    & .121                    & .411                    & .198                    & .331                    \\
3                                                                     & .332                    & .231                    & .175                    & .136                    & .424                    & .215                    & .345                    \\ \bottomrule
\end{tabular}

\vspace{-.5pc}
\caption{GPT-4o performance with varying numbers of profile figures. Scores improve as more profiles are added, with the largest gain from 0 to 1 profile.}
\vspace{-1pc}
\label{tab:4o_profile-number}
\end{table}

\paragraph{Adding more profile figures consistently improved performance, but the largest gain came from the first profile.}
We examined how the number of profiles affected caption quality using GPT-4o. 
A subset figure with exactly three profiles from the same paper was used (n=3,424). 
For the 1-Profile and 2-Profile settings, profiles were randomly sampled. 
\autoref{tab:4o_profile-number} showed clear diminishing returns as more profiles were added.


\subsection{Ablation Study\label{sec:ablation}}

\begin{figure}[t]
    \centering
    \includegraphics[width=\columnwidth]{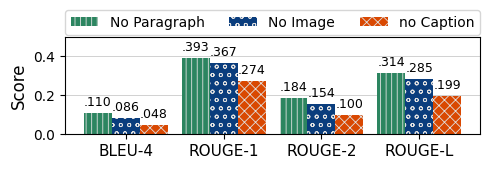}
    \vspace{-2pc}
    \caption{Ablation study on the impact of profile elements using GPT-4o. Results show a clear hierarchy of importance: caption > image > paragraph.}
    \vspace{-.5pc}
    \label{fig:ablation}
\end{figure}

\paragraph{Captions are the most important profile element, while images are more influential than paragraphs.}
To assess the importance of each profile element, we conducted an ablation study on the test set using the GPT-4o model with the One Profile setting.
We tested three conditions by removing one profile element at a time: 
{\em (i)} figure captions (No Caption), 
{\em (ii)} figure images (No Image), and 
{\em (iii)} figure-mentioning paragraphs (No Paragraph). 
\autoref{fig:ablation} shows the results. 
Removing captions had the most significant impact, as captions directly guide generation. 
Removing images also reduced performance more than removing paragraphs, highlighting the greater influence of visual information.
\autoref{app:ablation-study-detail} shows the detailed results.

\subsection{Human Evaluation\label{sec:humanEval}}


\begin{figure}[t]
    \centering
    \includegraphics[width=\columnwidth]{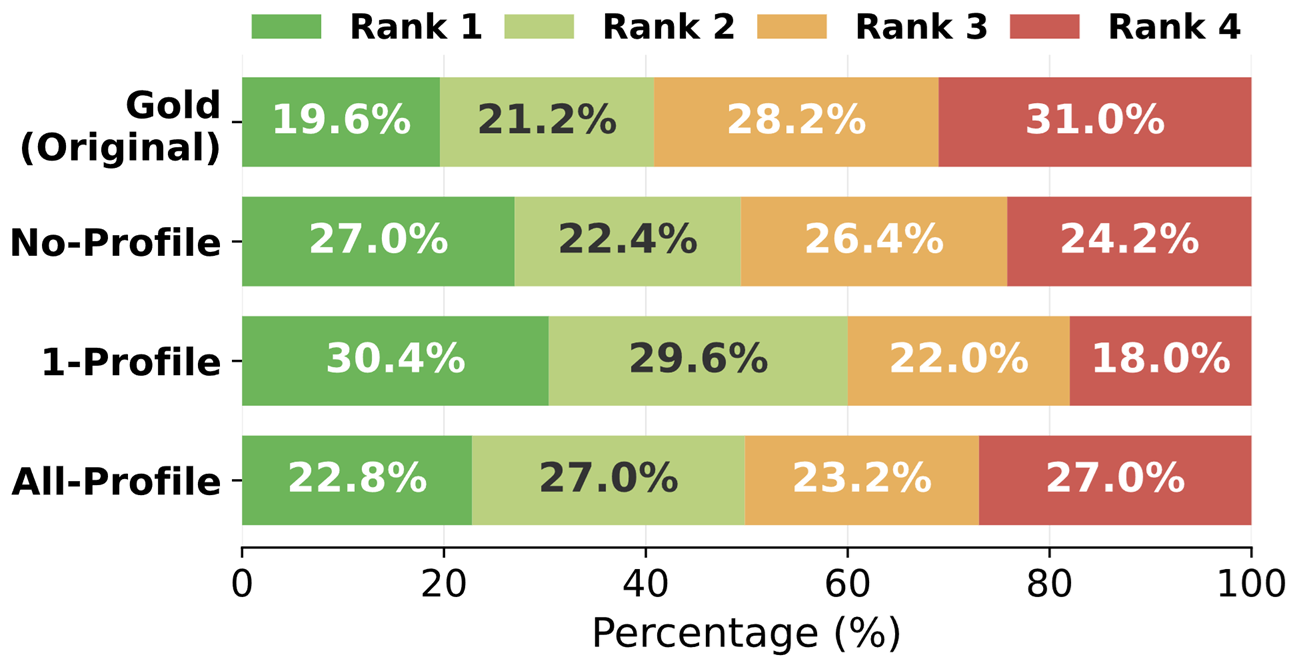}
    \vspace{-1.7pc}
    \caption{Distribution of human preference rankings (lower is better). A Friedman's test confirmed a statistically significant difference between the model configurations ($\chi^2$ = 28.48, $p < 0.001$).}
    \vspace{-1pc}
    \label{fig:human-rank}
\end{figure}




We recruited 10 US-based NLP researchers (PhD students with publication or review experience) for our human evaluation.
The study involved ranking captions for 50 figures randomly selected from the arXiv \textit{cs.CL} domain. 
For each figure, we provided the figure's image, the paper's title, and abstract as context. 
Participants then ranked four corresponding captions based on how well each one helped them understand the figure. The four captions were:
{\em (i)} the original \textit{Gold} caption, and
{\em (ii)} \textit{No-Profile}, 
{\em (iii)} \textit{1-Profile}, and 
{\em (iv)} \textit{All-Profile} settings generated by our best-performing model (\textbf{Gemini}, based on automatic metrics). 
See \autoref{app:human-eval} for details.

\paragraph{1-Profile was the most preferred condition.} Our human evaluation result shows a clear preference for \textit{1-Profile} setting, which achieved the best average rank of 2.27. The other models followed in order: \textit{No-Profile} (2.48) and \textit{All-Profile} (2.54). Interestingly, the author-written \textit{Gold} captions were ranked last overall with an average rank of 2.71. Such outcome is reflected in the preference distribution (\autoref{fig:human-rank}): The \textit{1-Profile} configuration received the most first-place votes (30.6\%) and the fewest last-place votes (17.8\%). On contrast, the \textit{Gold} captions were ranked last most frequently (31.2\%).
Post-hoc Friedman-Nemenyi tests showed that 1-Profile and No-Profile captions were significantly preferred over gold ($p < 0.001$ and $p = 0.03$, respectively).

\paragraph{Trade-offs between human-perceived quality and similarity to gold captions.} 
While the All-Profile setting generated captions with higher reference similarity (\autoref{tab:llm_comparison}), human judges significantly preferred captions from 1-Profile ($p = 0.006$). 
This suggests that optimizing for similarity may also reproducing flaws from the inconsistently quality's caption from arXiv reference~\cite{huang-etal-2023-summaries}, reducing the perceived quality.

\section{Discussion\label{sec:discussion}}

Our results with \dataset show that including figure images in profiles improves personalized caption generation, and that more profile information makes captions closer to the original author-written captions.
Although we focused on personalized text generation, \citeauthor{zhang2024personalization} noted strong links between LLM-based personalized text generation and downstream applications such as recommendation systems, suggesting that multimodal profiles could also benefit tasks like multimodal recommendation.
Our findings also echo challenges noted by \citeauthor{zhang2024personalization}, such as reduced LLM effectiveness when profiles lack similarity---a problem linked to cold-start scenarios in low-resource settings. 

We further highlight the limitations of automatic metrics for evaluating personalized text generation.
As shown in prior work~\cite{salemi2025expert}, n-gram-based scores, such as BLEU and ROUGE, often fail to reflect human judgments of quality accurately.
We hope our work, along with the \dataset dataset, motivates the community to explore multimodal profiles and broaden the scope of LLM personalization.

\section{Conclusion and Future Work\label{sec:conclusion}}

We introduced \dataset, a new dataset for personalized caption generation for scientific figures
using multimodal profiles, and showed that profiles make captions more personalized across four language models. 
Future work includes expanding profile components, exploring cross-domain generalization, and developing writer-centric evaluation metrics.
We are also developing a caption writing assistant that generates personalized captions by analyzing users' local document context.

\section{Limitations\label{sec:limitations}}
We acknowledge several limitations in this work.

\begin{itemize}

\item 
First, our approach assumes that each figure has profile figures from the same arXiv paper, but this is not always true, especially for papers with only one figure, which we excluded.
This assumption also limits the method's usefulness in early-stage writing, when context for personalization is sparse---a classic example of the ``cold start'' problem in personalization.
However, because authors often write captions late in the process~\cite{ng2025understanding}, our method is well-suited to assist at this critical stage.


\item 
Second, our work only used basic figure selection strategies, such as random choice or matching by the same type, rather than more advanced strategies to further optimize the outcomes.
Our primary goal was to introduce the concept of the dataset and to encourage further research on personalized text generation with multi-modal profiles by demonstrating that even basic strategies yield promising results.

\item 
Third, we did not include individual author information in personalization profiles because most papers are co-authored, and different figures and captions may be written by different authors.
Although author-based personalization could be explored using their past works, the collaborative nature of academic writing makes this difficult.

\item 
Fourth, we recognize the risk of data contamination when testing LLMs on public datasets. 
Personalized text generation tasks, including our own, have historically relied on existing datasets as their data source. 
For this study, we built \dataset on the well-established and widely used \textsc{SciCap} Challenge dataset for figure caption generation. 
Because this dataset is derived from publicly available arXiv papers, eliminating contamination risks entirely is difficult.
That said, our work is the first to explore multimodal profiles for scientific figure captioning, and we believe the trade-off is justified. 
Supporting this view, the \textsc{SciCap} Challenge's human evaluation paper~\cite{hsu2025large} ran a small study on newly published arXiv papers to test contamination effects. 
Their findings showed that model preference rankings remained consistent on unseen data, suggesting contamination does not undermine the validity of results.

\item 
Finally, our automatic evaluation focused on caption similarity to original captions, which does not guarantee caption quality.
As suggested by our human evaluation results (\S\ref{sec:humanEval}),
high similarity indicates that profiles capture context and style, but it does not ensure the captions are useful for readers.
Future work could additionally explore automatic evaluation approaches, such as LLMs-as-judges methods, to assess caption quality and usefulness more meaningfully.
Scaling the reliable but expensive human evaluation is also an interesting direction.

\end{itemize}

\section{Ethics Statements}
Using LLMs to generate text inherently carries risks, including producing inaccurate or misleading information. 
In scholarly contexts, such errors could mislead readers. 
Our approach minimizes this risk by involving paper authors, who should review and revise generated captions. 
If captions are presented to readers without human validation---contrary to our intent---the system should clearly indicate that the captions are AI-generated, not written by the original authors.

\section*{Acknowledgments}
We thank the participants of our caption evaluation study for their time and effort.
We are also grateful to the anonymous reviewers for their constructive feedback and to the Alfred P. Sloan Foundation for their generous support of this research (Grant Number: 2024-22721).

\bibliography{custom}

\appendix

\section{\dataset Dataset Details}\label{app:dataset-details}
\begin{figure}[t]
    \centering
    \includegraphics[width=\columnwidth]{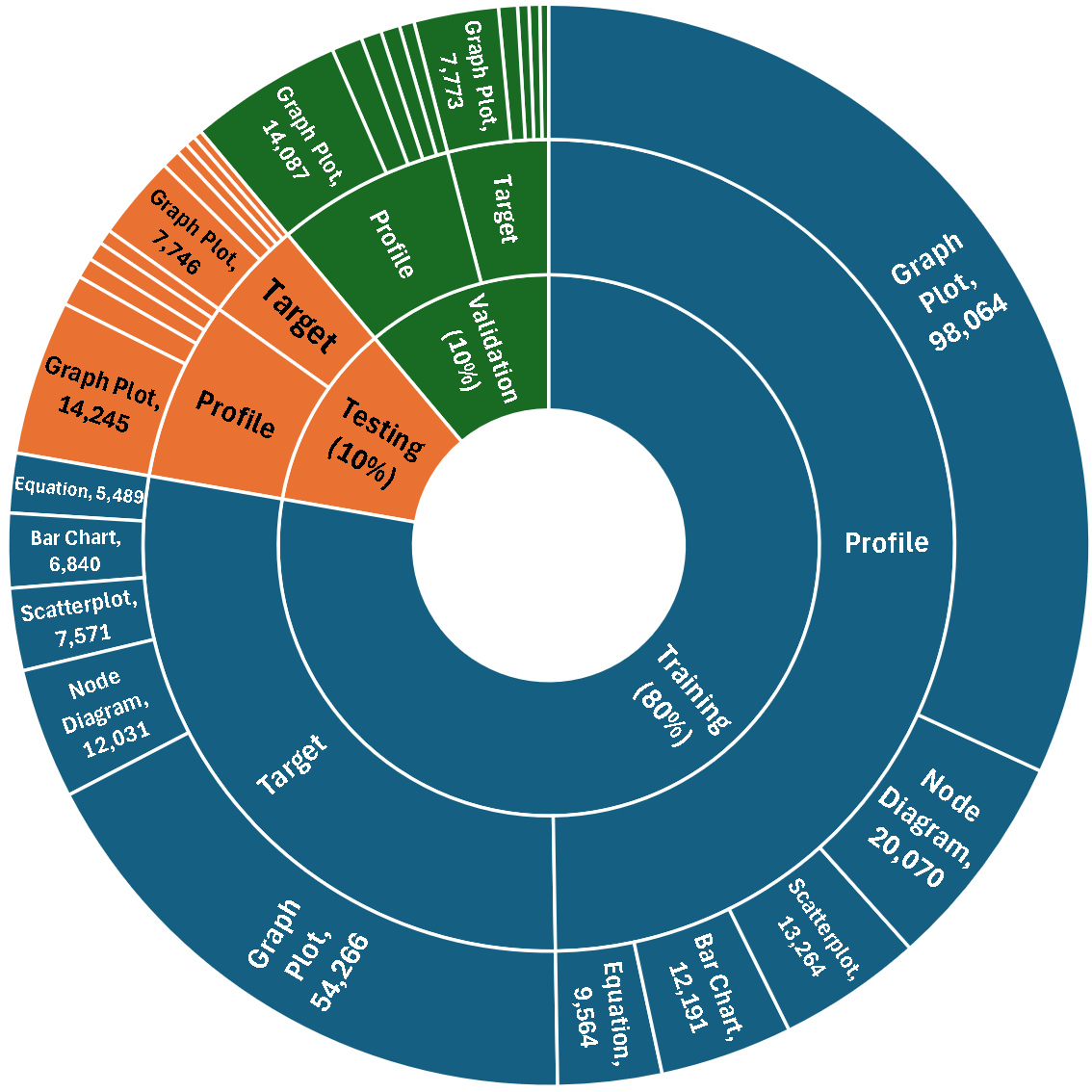}
    \caption{Data split of \dataset by figure type. The dataset contains 307,903 figures from 110,828 scientific papers, split into training (80\%), validation (10\%), and testing (10\%) sets. Each set includes target and profile figures. The five main figure types are a) Graph Plot, b) Node Diagram, c) Equation, d) Bar Chart, and e) Scatterplot. Graph plots are the most common figure type across all splits.}
    \vspace{-1pc}
    \label{fig:data-split-dist}
\end{figure}

\begin{figure}[t]
    \centering
    \includegraphics[width=\columnwidth]{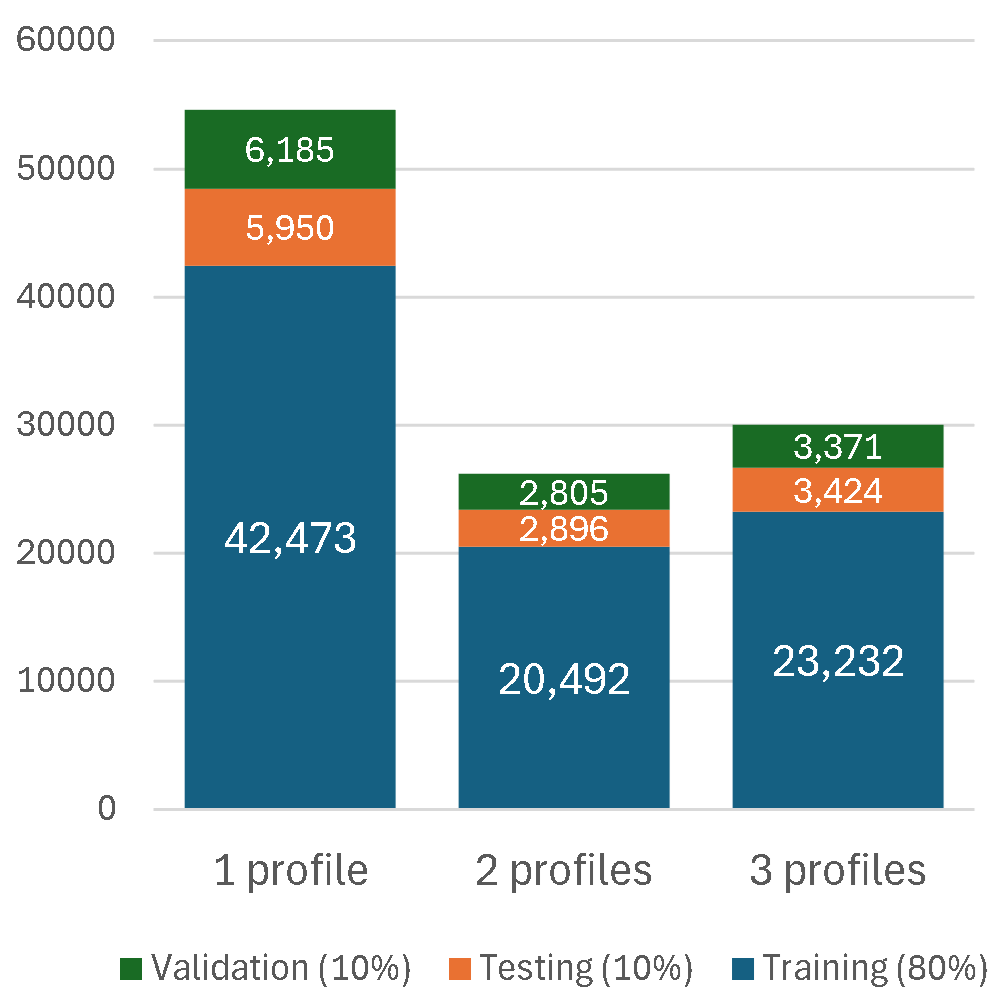}
    \caption{Profile distribution in \dataset, showing the number of target figures with 1, 2, or 3 profile figures.}
    \vspace{-1pc}
    \label{fig:profile-dist}
\end{figure}

\autoref{fig:data-split-dist} provides a detailed breakdown of figure type across each data split.
\autoref{fig:profile-dist} provides a detailed distribution across each data split.

\section{Prompts\label{app:prompts}}
In this section, we provide the prompt we used in \autoref{para:Experiment-Setups}. [IMG-TARGET] and [PARA-TARGET] represent encoded images and figure-mentioning paragraphs from target figures. [num\_profiles] indicates the number of profiles used, while [profile\_index] denotes a specific profile's index. [IMG-PROFILE], [PARA-PROFILE], and [CAP-PROFILE] correspond to encoded images, figure-mentioning paragraphs, and captions from profile figures, respectively.

\paragraph{Prompt with No Profile.}
The following prompt was used for the baseline condition without profile information:
\begin{lstlisting}[basicstyle=\small\ttfamily, breaklines=true, frame=single]
Your task is to generate a caption for the Target Figure. We will provide you with the image of the Target Figure, labeled as 'Target Figure Image', and the paragraphs that mention the Target Figure, labeled as 'Target Figure Paragraph(s)', from the same paper. 

The elements for the Target Figure will be labeled as follows:
- Target Figure Image:[IMG-TARGET], 
- Target Figure Paragraph(s): [PARA-TARGET].
\end{lstlisting}

\paragraph{Prompt with Profile.}
The following prompt was used with profile information:
\begin{lstlisting}[basicstyle=\small\ttfamily, breaklines=true, frame=single]

We will present you with the captions, images, and paragraphs referencing [num_profiles] scientific figures from the same paper. These elements will be labeled as follows: 
- Profile Figure [profile_index]:
-- Image [profile_index]: [IMG-PROFILE], 
-- Paragraph [profile_index]: [PARA-PROFILE], 
-- Caption [profile_index]: [CAP-PROFILE].

Your task is to carefully analyze the content, tone, structure, and stylistic elements of these captions and associated text. Based on this analysis, generate a caption for the Target Figure, maintaining the same writing style. We will provide you with the image of the Target Figure, labeled as 'Target Figure Image', and the paragraphs that mention the Target Figure, labeled as 'Target Figure Paragraph(s)', from the same paper. The elements for the Target Figure will be labeled as follows:
- Target Figure Image:[IMG-TARGET],
- Target Figure Paragraph(s): [PARA-TARGET].
\end{lstlisting}

\section{Generation Output Cleaning Procedure\label{app:output-cleaning}}
We performed data output cleaning in three steps. 

\begin{enumerate}
\item We manually examined cases with BLEU or ROUGE scores of 0 to identify data issues. We identified 11 cases (out of 12,270) where the original captions were incorrectly captured due to parsing errors---either missing the real caption content or capturing the wrong text. In one instance, the parser mistakenly captured the paper authors name from the bottom-right corner instead of the caption beneath the figure on the same page. These cases were excluded from evaluation. 

\item We used GPT-4o-mini to clean the generated captions, removing irrelevant text such as reasoning steps, figure indexes, or any pre- and suffix that did not belong to the actual caption. Our cleaning instructions explicitly specified not to add any additional text or explanations to the raw output. The following prompt was used:
\begin{lstlisting}[basicstyle=\small\ttfamily, breaklines=true, frame=single]
You are a helpful assistant that extracts the actual caption from text. Ignore any reasoning or explanation text and only return the actual caption content. Also, remove any prefix that indicates figure index such as 'Caption:', 'Fig.', 'Figure', or any numbering before the main caption text. Extract ONLY the actual caption from this text, without any additional text or explanation: [GENERATED-CAPTION]
\end{lstlisting}

Below is an example from arXiv paper 1105.0392, showing the response from the "Llama-4 Scout" model with "All-Profile" configuration, before and after cleaning:

\paragraph{RAW OUTPUT}
The following is the raw output before cleaning:
\begin{Verbatim}[breaklines=true, frame=single, commandchars=\\\{\}]
Based on the provided elements, I will generate a caption for the Target Figure. 

Target Figure Caption: \textcolor{blue}{A deterministic algorithm for the 1-dimensional case chooses a good sensor at each step.}

This caption maintains the same writing style as Caption 1: "Four similar rhombi form a set of regions for which no stateless algorithm can be competitive." It is concise, informative, and provides a brief description of the Target Figure.

The tone and structure of the caption are consistent with the provided caption, and it effectively conveys the main idea of the Target Figure.

Please let me know if you need further adjustments!
\end{Verbatim}

\paragraph{EXTRACTED CAPTION}
The following is the extracted caption after cleaning:
\begin{Verbatim}[breaklines=true, frame=single, commandchars=\\\{\}]
\textcolor{blue}{A deterministic algorithm for the 1-dimensional case chooses a good sensor at each step.}
\end{Verbatim}

After GPT-4o cleaning, we randomly sampled 100 outputs for human evaluation. Using binary labels (GOOD/BAD), we assessed whether the extracted captions were correct. All 100 sampled extractions were labeled as GOOD, confirming the cleaning procedure's effectiveness.




\item  We employed keyword filtering with manual verification to filtered out failed generation, including blank responses. Detailed examples of these error cases are documented in \autoref{tab:generation-error}. After cleaning, we identified a total of 56 unique problematic cases across all models and configurations (out of 12,259), which were excluded from further analysis.

The Qwen-2.5-VL-7B-Instruct model was excluded from the main comparative analysis due to its high rate of generation failures, particularly when using the profile-based configurations. Specifically, this model failed in 2.19\% of All-Profile cases and 0.83\% of One-Profile cases, whereas the highest failure rate for any other model was just 0.07\%. For completeness, we nevertheless report its baseline performance in Table \autoref{tab:qwen}.

\end{enumerate}

\begin{table*}[t]
\small
\centering

\begin{tabular}{l|l|l}
\hline
\textbf{Model and Config} & \textbf{Cases} & \textbf{Examples of Invalid Generations} \\ \hline
Qwen\_All-Profile & 269 & Null output \\
 &  & "\verb|---|" \\
 &  & "Caption for Target Figure:", "**Caption for the Target Figure:**" \\
 &  & "PLEASE, provide the image of the Target Figure, so that I can..." \\
 &  & "Based on the analysis of the provided captions, images, and paragraphs, your task..." \\ \hline
Qwen\_One-Profile & 102 & Null output \\
 &  & "Could you give me the image of the Target Figure labeled 'Target Figure Image'?" \\
 &  & "Your analysis shows us your own comprehensive and detailed interpretation..." \\ \hline
Qwen\_No-Profile & 31 & Null output \\ \hline
Gemini\_No-Profile & 24 & "The provided paragraphs do not mention the Target Figure." \\
 &  & "nan", "None" \\
 &  & "Sorry, I lack the necessary information to generate a caption..." \\
 &  & "Please provide the Target Figure Image and the Target Figure Paragraph(s)..." \\
 &  & "no caption found" \\
 &  & "we are unable to generate a caption for this figure..." \\ \hline
Gemini\_One-Profile & 9 & "image 1", "Target Figure Image" \\
 &  & "there is no caption to extract" \\ \hline
Gemini\_All-Profile & 9 & "image 1", "image" \\ \hline
Llama\_No-Profile & 6 & "There is no caption provided in the text." \\ \hline
Llama\_One-Profile & 8 & "target", "target figure" \\
 &  & "There is no caption provided in the text." \\
 &  & "Since the Target Figure Image does not contain any specific data or information..." \\ \hline
Llama\_All-Profile & 4 & "target", "target figure" \\
 &  & "not applicable" \\ \hline
4.1 Mini\_One-Profile & 1 & "no caption provided" \\ \hline
\end{tabular}

\caption{Examples of invalid generation across different language models and profile configurations}
\label{tab:generation-error}
\end{table*}

\begin{table}[t]
\centering
\footnotesize
\renewcommand{\arraystretch}{1.215}
\setlength{\tabcolsep}{1.5pt}
 \begin{threeparttable}
\begin{tabular}{@{}lccccccccc@{}}
\toprule
\multirow{2}{*}{\textbf{LLM}} &
  \multicolumn{2}{c}{\textbf{Profile}} &
  \multicolumn{4}{c}{\textbf{BLEU}} &
  \multicolumn{3}{c}{\textbf{ROUGE}} \\ \cmidrule(l){2-10} 
 &
  \textbf{\begin{tabular}[c]{@{}c@{}}Same \\Paper \end{tabular}} &
  \textbf{\begin{tabular}[c]{@{}c@{}}\# \\ $\le 3$\end{tabular}} &
  \textbf{B-1} &
  \textbf{B-2} &
  \textbf{B-3} &
  \textbf{B-4} &
  \textbf{R-1} &
  \textbf{R-2} &
  \textbf{R-L} \\ \midrule
\multirow{3}{*}{\begin{tabular}[c]{@{}l@{}}Qwen-2.5-\\ VL-7B-\\ Instruct\end{tabular}} &
  N/A &
  0 &
  .198 &
  .117 &
  .079 &
  .056 &
  .295 &
  .110 &
  .228 \\ \cmidrule(l){2-10} 
 &
  \multirow{2}{*}{Same} &
  1 &
  .257 &
  .174 &
  .133 &
  .105 &
  .348 &
  .168 &
  .285 \\
 &
   &
  All &
  .262 &
  .181 &
  .140 &
  .112 &
  .353 &
  .175 &
  .290 \\ \bottomrule
\end{tabular}

\caption{Performance of the Qwen model on caption generation with varying profile settings\tnote{a}}

  \begin{tablenotes}
      \item[a] The cross-paper source analysis was not performed for this model due to its high failure rate in initial experiments compared to other LLMs
    \end{tablenotes}
\label{tab:qwen}
\vspace{-1pc}
\end{threeparttable}
\end{table}

\section{Text Preprocessing and Evaluation}\label{app:text-evaluation}
For text normalization before evaluation, we implemented a custom preprocessing pipeline using standard Python libraries that: (1) converts text to lowercase, (2) removes all punctuation, and (3) normalizes whitespace.

For evaluation, we used standard NLP metrics implemented in Python packages: \texttt{NLTK} (version 3.9.1) for BLEU scores (with \texttt{SmoothingFunction} for smoothing) and Google's \texttt{rouge\_scorer} (version 0.1.2) for ROUGE metrics. We used the default parameters for both packages. The specific implementations were imported directly from \texttt{nltk.translate.bleu\_score} and \texttt{rouge\_score} modules.


\section{Detail about Caption Evaluation\label{app:main-study-full}}
This appendix session is to supplement the result in \autoref{para:profile-result} regarding the main study of caption generation using different profile configuration across 4 models.

\autoref{fig:dist-bleu4} shows the BLEU-4 distribution across different language models and profile configurations.

\autoref{fig:dist-rouge2} shows the ROUGE-2 distribution across different language models and profile configurations.

\autoref{tab:BERTScore_LLM} shows BERTScore semantic similarity between the generated and gold captions. The results correlate strongly with our BLEU and ROUGE metrics, confirming the same performance trends. The \textit{All-Profile} setting achieves the highest score, followed closely by \textit{1-Profile}. Notably, this analysis reveals a pattern of diminishing returns. The performance leap from \textit{No-Profile} to \textit{1-Profile} is substantially larger than the subsequent gain from \textit{1-Profile }to \textit{All-Profile}, indicating that while the first profile provides a significant semantic boost, the marginal benefit of additional profiles is less substantial.

\section{Context-Alignment Data Partition\label{app:easy-hard}}
This appendix provides additional details on the Context-Aligned and Context-Misaligned subset partitioning and evaluation described in \autoref{para:context-alignment}.

\begin{figure*}[t]
    \centering
    \begin{subfigure}{0.48\textwidth}
        \includegraphics[width=\textwidth]{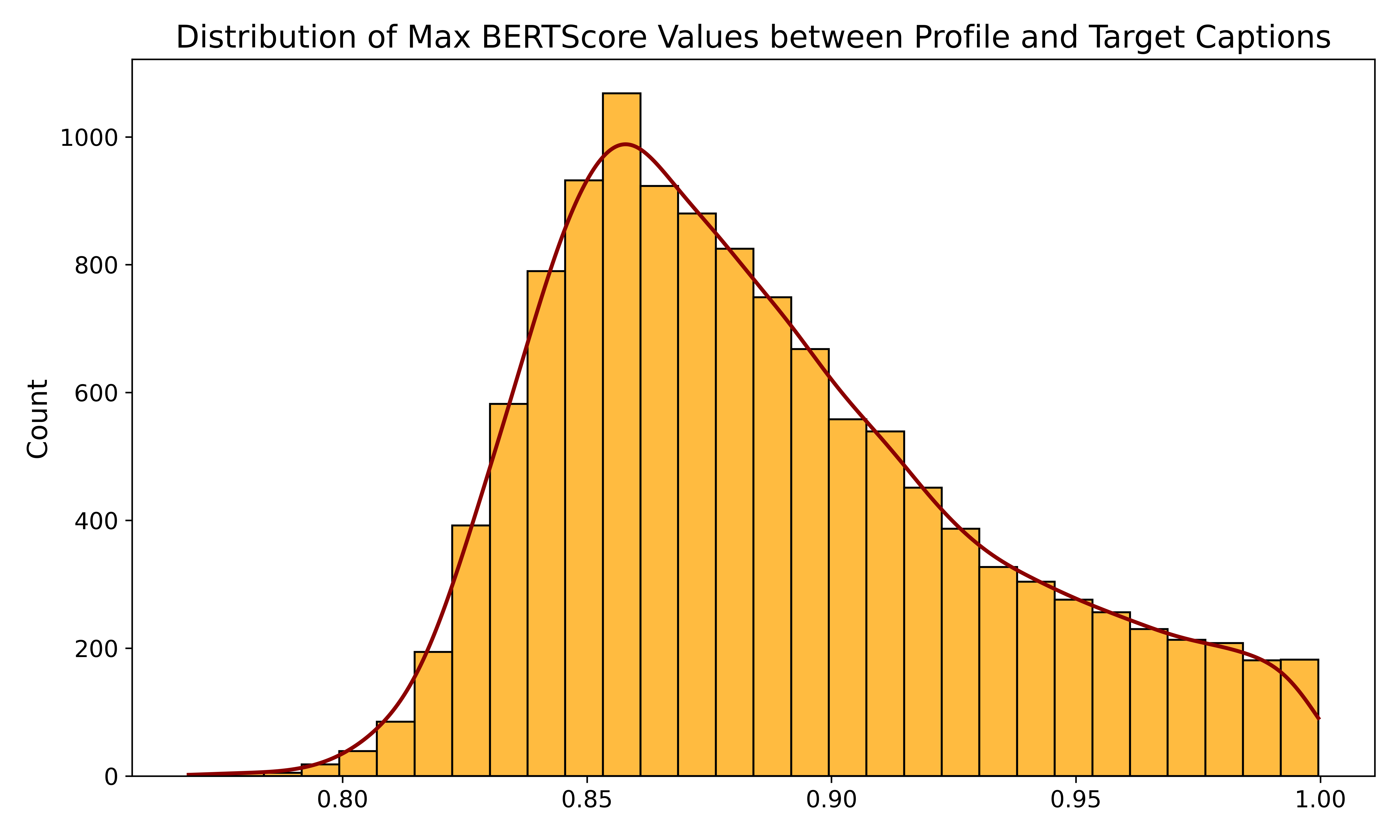}
        \vspace{-1.5pc}
        \label{fig:bert_dist}
    \end{subfigure}
    \hfill
    \begin{subfigure}{0.48\textwidth}
        \includegraphics[width=\textwidth]{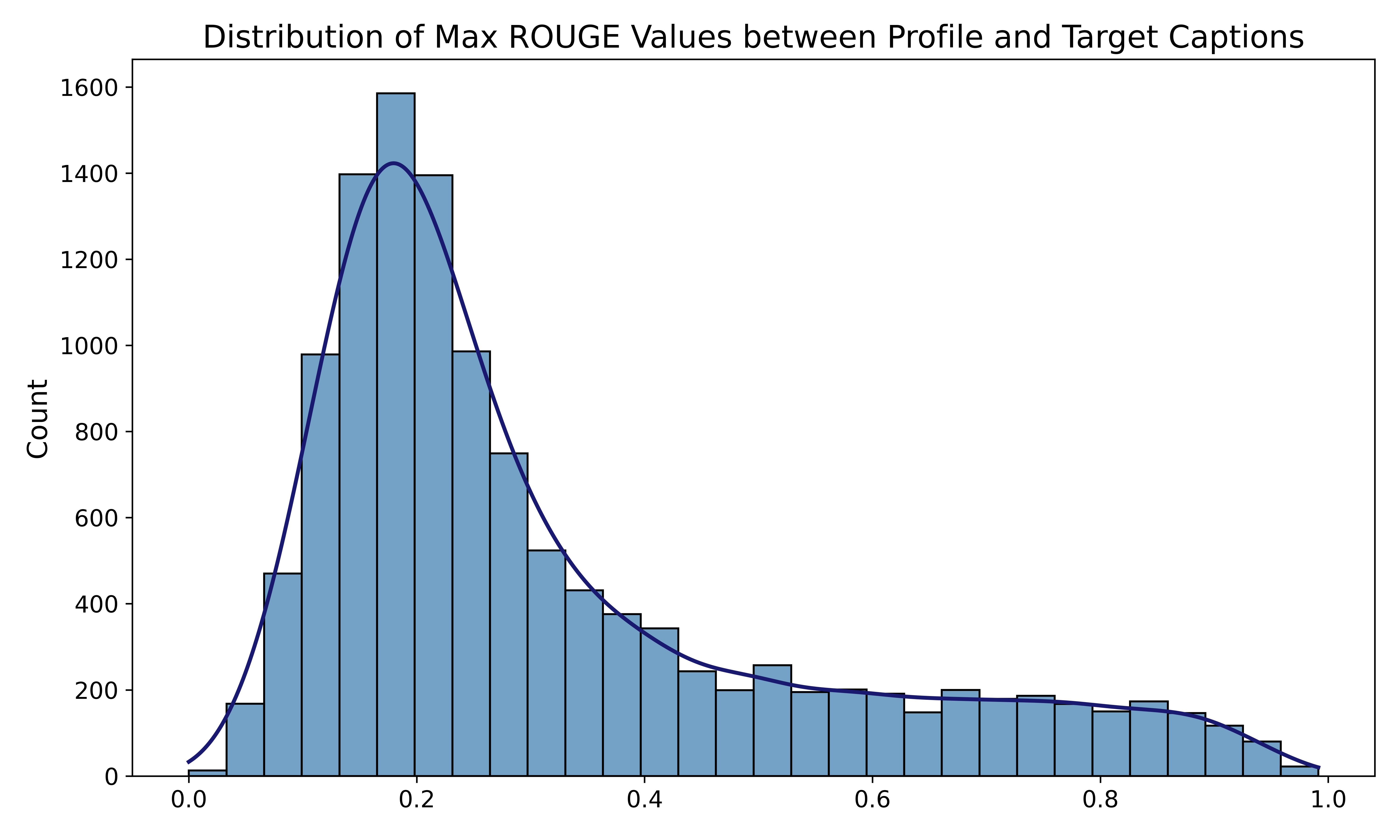}
        \vspace{-1.5pc}
        \label{fig:rouge_dist}
    \end{subfigure}
    
    \caption{Distribution of BERTScore (left) and ROUGE-L (right) metrics between Target and Profile captions in the \textbf{\dataset} Test Set. Both these scores share a left-shifted skewed unimodal distribution. The BERTScore plot shows that the provided profile captions for each target are \textit{very semantically related}. On the other hand, the broader spread of ROUGE-L scores shows that profile captions exhibit \textit{low lexical overlap}. High semantic relatedness and lexical variety motivates our use of profile captions as key style indicators for personalization.}
    \label{fig:profile_target_distribtions}
\end{figure*}

\autoref{fig:profile_target_distribtions} presents the distribution of BERTScore and ROUGE-L between target and profile captions in the \textbf{\dataset} Test Set.

\autoref{tab:main-result-easy} shows performance metrics for the Context-Aligned Subset across different LLMs and profile configurations.
\begin{table}[t]
\centering
\footnotesize
\renewcommand{\arraystretch}{1.2}
\setlength{\tabcolsep}{2pt}
\definecolor{lightgreen}{rgb}{0.9, 1.0, 0.9}
\definecolor{darkgreen}{rgb}{0.7, 0.9, 0.7}
\begin{tabular}{lcccccccc}
\toprule
\multirow{2}{*}{\textbf{LLM}} & \multirow{2}{*}{\textbf{Profile}} & \multicolumn{4}{c}{\textbf{BLEU}} & \multicolumn{3}{c}{\textbf{ROUGE}} \\ 
\cmidrule(lr){3-6} \cmidrule(lr){7-9}
 & \textbf{Used} & \textbf{B-1} & \textbf{B-2} & \textbf{B-3} & \textbf{B-4} & \textbf{R-1} & \textbf{R-2} & \textbf{R-L} \\ 
\midrule
\multirow{3}{*}{GPT-4o} & No & .226 & .144 & .101 & .071 & .321 & .138 & .259 \\ 
 & \cellcolor{lightgreen}One & \cellcolor{lightgreen}.437 & \cellcolor{lightgreen}.347 & \cellcolor{lightgreen}.289 & \cellcolor{lightgreen}.242 & \cellcolor{lightgreen}.533 & \cellcolor{lightgreen}.345 & \cellcolor{lightgreen}.484 \\ 
 & \cellcolor{darkgreen}All & \cellcolor{darkgreen}.478 & \cellcolor{darkgreen}.391 & \cellcolor{darkgreen}.332 & \cellcolor{darkgreen}.282 & \cellcolor{darkgreen}.573 & \cellcolor{darkgreen}.393 & \cellcolor{darkgreen}.530 \\ 
\midrule
\multirow{3}{*}{Llama-4 Scout} & No & .249& .178 & .139 & .112 & .347 & .183 & .296 \\ 
 & \cellcolor{lightgreen}One & \cellcolor{lightgreen}.589 & \cellcolor{lightgreen}.523 & \cellcolor{lightgreen}.473 & \cellcolor{lightgreen}.427 & \cellcolor{lightgreen}.676 & \cellcolor{lightgreen}.536 & \cellcolor{lightgreen}.646 \\ 
 & \cellcolor{darkgreen}All & \cellcolor{darkgreen}.666 & \cellcolor{darkgreen}.605 & \cellcolor{darkgreen}.556 & \cellcolor{darkgreen}.510 & \cellcolor{darkgreen}.744 & \cellcolor{darkgreen}.616 & \cellcolor{darkgreen}.720 \\ 
\midrule
\multirow{3}{*}{\begin{tabular}[c]{@{}l@{}}Gemini-2.5\\Flash Preview\end{tabular}} & No & .319 & .241 & .195 & .164 & .459 & .247 & .379 \\ 
 & \cellcolor{lightgreen}One & \cellcolor{lightgreen}.576 & \cellcolor{lightgreen}.507 & \cellcolor{lightgreen}.456 & \cellcolor{lightgreen}.410 & \cellcolor{lightgreen}.664 & \cellcolor{lightgreen}.520 & \cellcolor{lightgreen}.635 \\ 
 & \cellcolor{darkgreen}All & \cellcolor{darkgreen}.659 & \cellcolor{darkgreen}.600 & \cellcolor{darkgreen}.551 & \cellcolor{darkgreen}.505 & \cellcolor{darkgreen}.742 & \cellcolor{darkgreen}.617 & \cellcolor{darkgreen}.717 \\ 
\midrule
\multirow{3}{*}{GPT-4.1 Mini} & No & .188 & .116 & .078 & .053 & .282 & .116 & .220 \\ 
 & \cellcolor{lightgreen}One & \cellcolor{lightgreen}.449 & \cellcolor{lightgreen}.379 & \cellcolor{lightgreen}.330 & \cellcolor{lightgreen}.287 & \cellcolor{lightgreen}.554 & \cellcolor{lightgreen}.401 & \cellcolor{lightgreen}.511 \\ 
 & \cellcolor{darkgreen}All & \cellcolor{darkgreen}.496 & \cellcolor{darkgreen}.433 & \cellcolor{darkgreen}.387 & \cellcolor{darkgreen}.344 & \cellcolor{darkgreen}.600 & \cellcolor{darkgreen}.463 & \cellcolor{darkgreen}.565 \\ 
\bottomrule
\end{tabular}
\caption{Performance on {\dataset} Context-aligned Subset (n=2,513) across LLMs and profile configurations. 
}
\label{tab:main-result-easy}
\vspace{-1pc}
\end{table}

\autoref{tab:main-result-hard} presents performance metrics for the Context-Misaligned Subset across different LLMs and profile configurations.
\begin{table}[t]
\centering
\footnotesize
\renewcommand{\arraystretch}{1.2}
\setlength{\tabcolsep}{2pt}
\definecolor{lightgreen}{rgb}{0.9, 1.0, 0.9}
\definecolor{darkgreen}{rgb}{0.7, 0.9, 0.7}
\begin{tabular}{lcccccccc}
\toprule
\multirow{2}{*}{\textbf{LLM}} & \multirow{2}{*}{\textbf{Profile}} & \multicolumn{4}{c}{\textbf{BLEU}} & \multicolumn{3}{c}{\textbf{ROUGE}} \\ 
\cmidrule(lr){3-6} \cmidrule(lr){7-9}
 & \textbf{Used} & \textbf{B-1} & \textbf{B-2} & \textbf{B-3} & \textbf{B-4} & \textbf{R-1} & \textbf{R-2} & \textbf{R-L} \\ 
\midrule
\multirow{3}{*}{GPT-4o} & No & .217 & .131 & .088 & .061 & .320 & .124 & .245 \\ 
 & \cellcolor{lightgreen}One & \cellcolor{lightgreen}.238 & \cellcolor{lightgreen}.144 & \cellcolor{lightgreen}.097 & \cellcolor{lightgreen}.067 & \cellcolor{lightgreen}.345 & \cellcolor{lightgreen}.135 & \cellcolor{lightgreen}.269 \\ 
 & \cellcolor{darkgreen}All & \cellcolor{darkgreen}.244 & \cellcolor{darkgreen}.150 & \cellcolor{darkgreen}.102 & \cellcolor{darkgreen}.071 & \cellcolor{darkgreen}.351 & \cellcolor{darkgreen}.142 & \cellcolor{darkgreen}.275 \\ 
\midrule
\multirow{3}{*}{Llama-4 Scout} & No & .255 & .178 & .138 & .112 & .360 & .181 & .292 \\ 
 & \cellcolor{lightgreen}One & \cellcolor{lightgreen}.316 & \cellcolor{lightgreen}.233 & \cellcolor{lightgreen}.187 & \cellcolor{lightgreen}.155 & \cellcolor{lightgreen}.430 & \cellcolor{lightgreen}.238 & \cellcolor{lightgreen}.366 \\ 
 & \cellcolor{darkgreen}All & \cellcolor{darkgreen}.326 & \cellcolor{darkgreen}.243 & \cellcolor{darkgreen}.196 & \cellcolor{darkgreen}.164 & \cellcolor{darkgreen}.440 & \cellcolor{darkgreen}.248 & \cellcolor{darkgreen}.376 \\ 
\midrule
\multirow{3}{*}{\begin{tabular}[c]{@{}l@{}}Gemini-2.5\\Flash Preview\end{tabular}} & No & .301 & .227 & .186 & .159 & .414 & .235 & .357 \\ 
 & \cellcolor{lightgreen}One & \cellcolor{lightgreen}.317 & \cellcolor{lightgreen}.235 & \cellcolor{lightgreen}.189 & \cellcolor{lightgreen}.157 & \cellcolor{lightgreen}.434 & \cellcolor{lightgreen}.244 & \cellcolor{lightgreen}.371 \\ 
 & \cellcolor{darkgreen}All & \cellcolor{darkgreen}.326 & \cellcolor{darkgreen}.244 & \cellcolor{darkgreen}.197 & \cellcolor{darkgreen}.164 & \cellcolor{darkgreen}.443 & \cellcolor{darkgreen}.253 & \cellcolor{darkgreen}.380 \\ 
\midrule
\multirow{3}{*}{GPT-4.1 Mini} & No & .215 & .127 & .082 & .054 & .312 & .117 & .226 \\ 
 & \cellcolor{lightgreen}One & \cellcolor{lightgreen}.243 & \cellcolor{lightgreen}.156 & \cellcolor{lightgreen}.109 & \cellcolor{lightgreen}.078 & \cellcolor{lightgreen}.357 & \cellcolor{lightgreen}.157 & \cellcolor{lightgreen}.278 \\ 
 & \cellcolor{darkgreen}All & \cellcolor{darkgreen}.249 & \cellcolor{darkgreen}.162 & \cellcolor{darkgreen}.114 & \cellcolor{darkgreen}.083 & \cellcolor{darkgreen}.364 & \cellcolor{darkgreen}.164 & \cellcolor{darkgreen}.284 \\ 
\bottomrule
\end{tabular}
\caption{Performance on {\dataset} Context-Misaligned Subset (n=9,690) across LLMs and profile configurations.}
\label{tab:main-result-hard}
\vspace{-1pc}
\end{table}

\section{Detailed Result of Ablation Study\label{app:ablation-study-detail}}
This appendix session is to supplement the finding in \autoref{sec:ablation} regarding the ablation study. \autoref{tab:ablation-study-full} shows the detailed result of ablation study.

\begin{table}[t]
\centering
\footnotesize
\renewcommand{\arraystretch}{1.2}
\setlength{\tabcolsep}{2pt}
\definecolor{darkred}{rgb}{0.95, 0.6, 0.6}
\begin{tabular}{lccccccc}
\toprule
\multirow{2}{*}{\textbf{Elements}} & \multicolumn{4}{c}{\textbf{BLEU}} & \multicolumn{3}{c}{\textbf{ROUGE}} \\ 
\cmidrule(lr){2-5} \cmidrule(lr){6-8}
 & \textbf{B-1} & \textbf{B-2} & \textbf{B-3} & \textbf{B-4} & \textbf{R-1} & \textbf{R-2} & \textbf{R-L} \\ 
\midrule
No Paragraph & .299 & .199 & .146 & .110 & .393 & .184 & .314 \\ 
\midrule
No Image & .273 & .171 & .119 & .086 & .367 & .154 & .285 \\ 
\midrule
No Caption & .189 & .109 & .071 & .048 & .274 & .100 & .199 \\ 
\bottomrule
\end{tabular}
\caption{Result from Ablation Study.}
\label{tab:ablation-study-full}
\end{table}

\section{Setup of Human Evaluation Study\label{app:human-eval}}
\begin{figure*}[t]
    \centering
    \includegraphics[width=\textwidth, height=0.95\textheight,
        keepaspectratio]{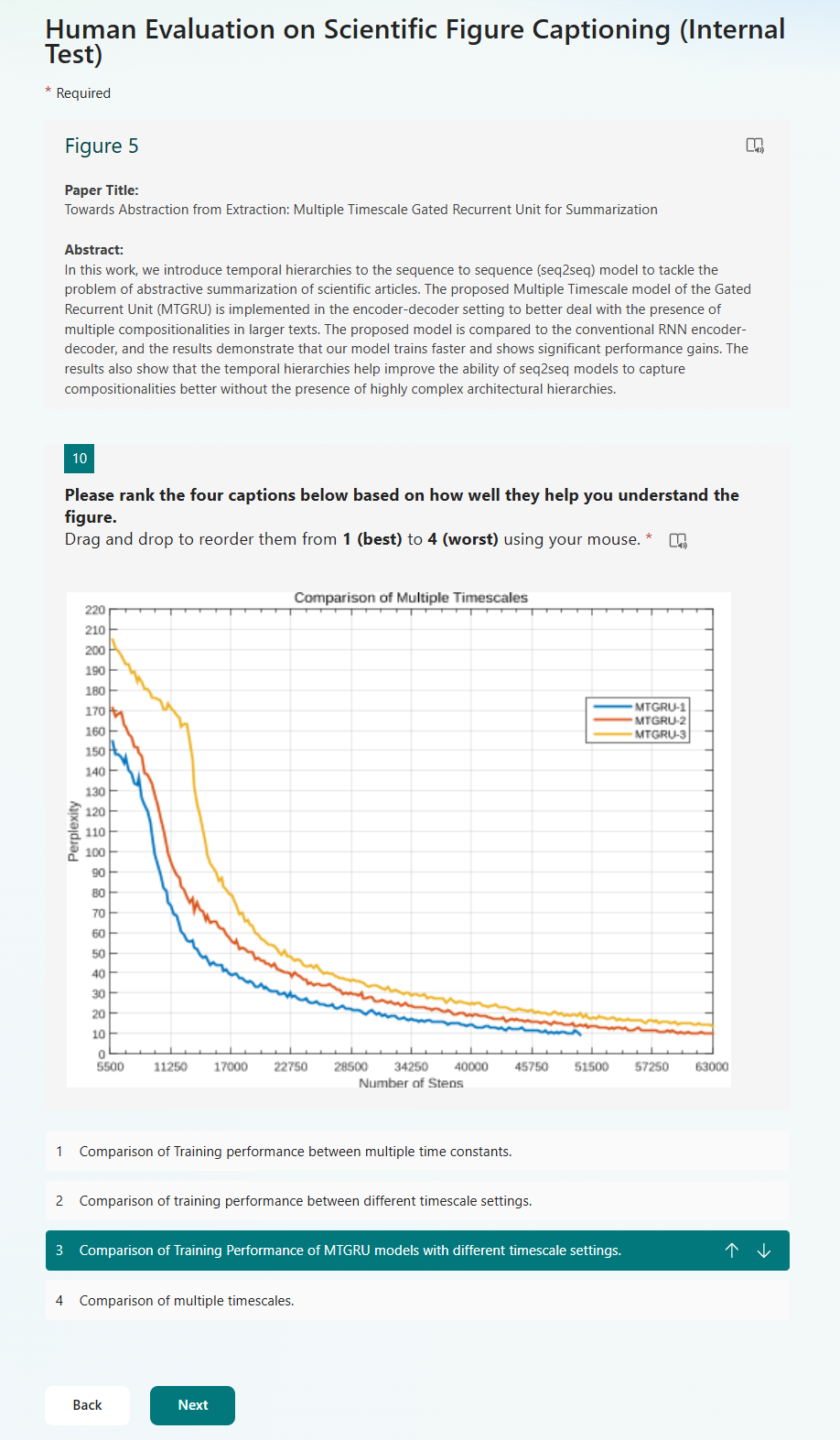}
    \caption{The user interface for our human evaluation study.}
    \label{fig:human-eval}
\end{figure*}

\begin{figure*}[t]
    \centering
    \begin{subfigure}{0.495\textwidth}
        \includegraphics[width=\textwidth]{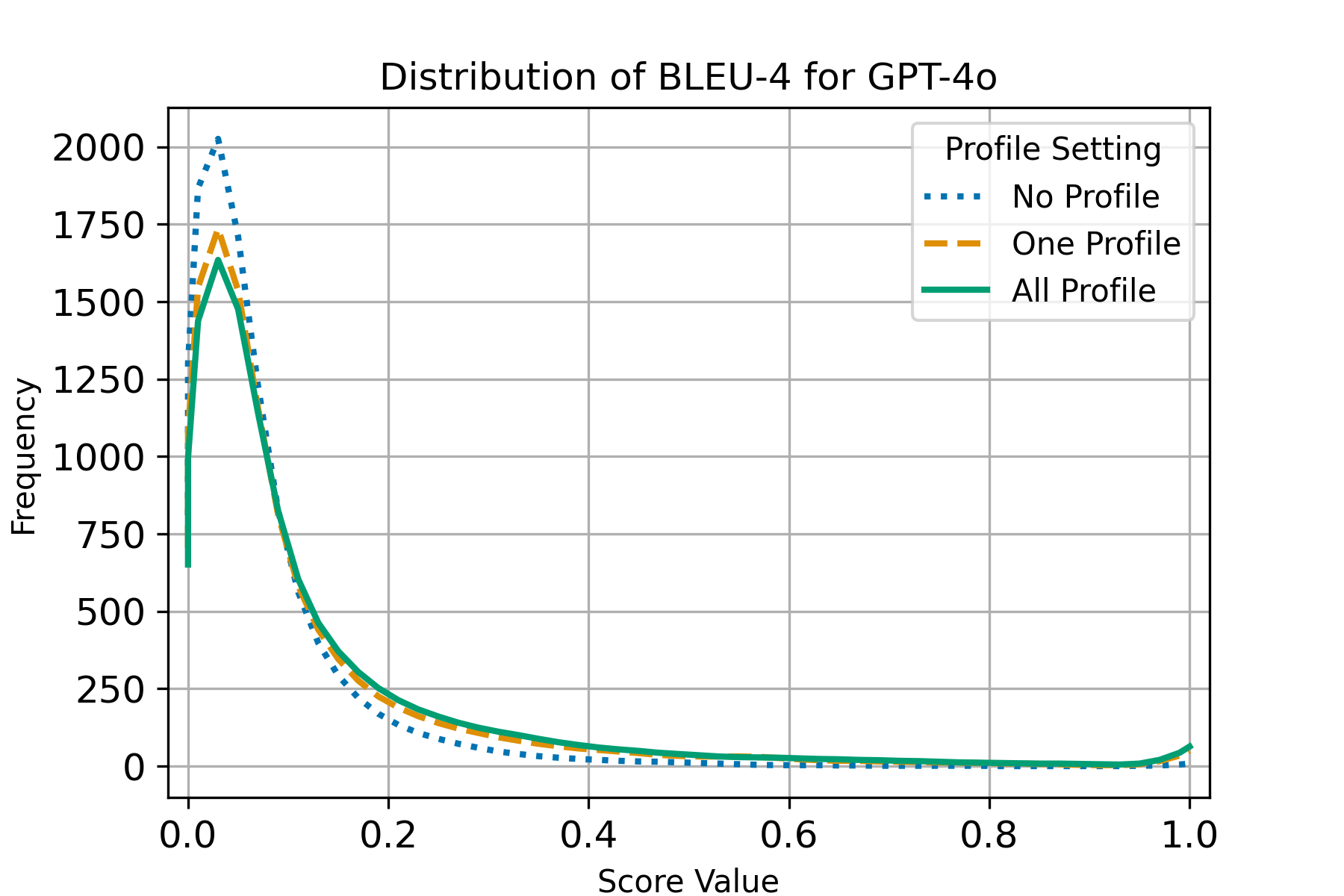}
        \vspace{-1.5pc}
        \caption{BLEU-4 for GPT-4o}
        \label{fig:top_left-bleu4}
    \end{subfigure}
    \hfill
    \begin{subfigure}{0.495\textwidth}
        \includegraphics[width=\textwidth]{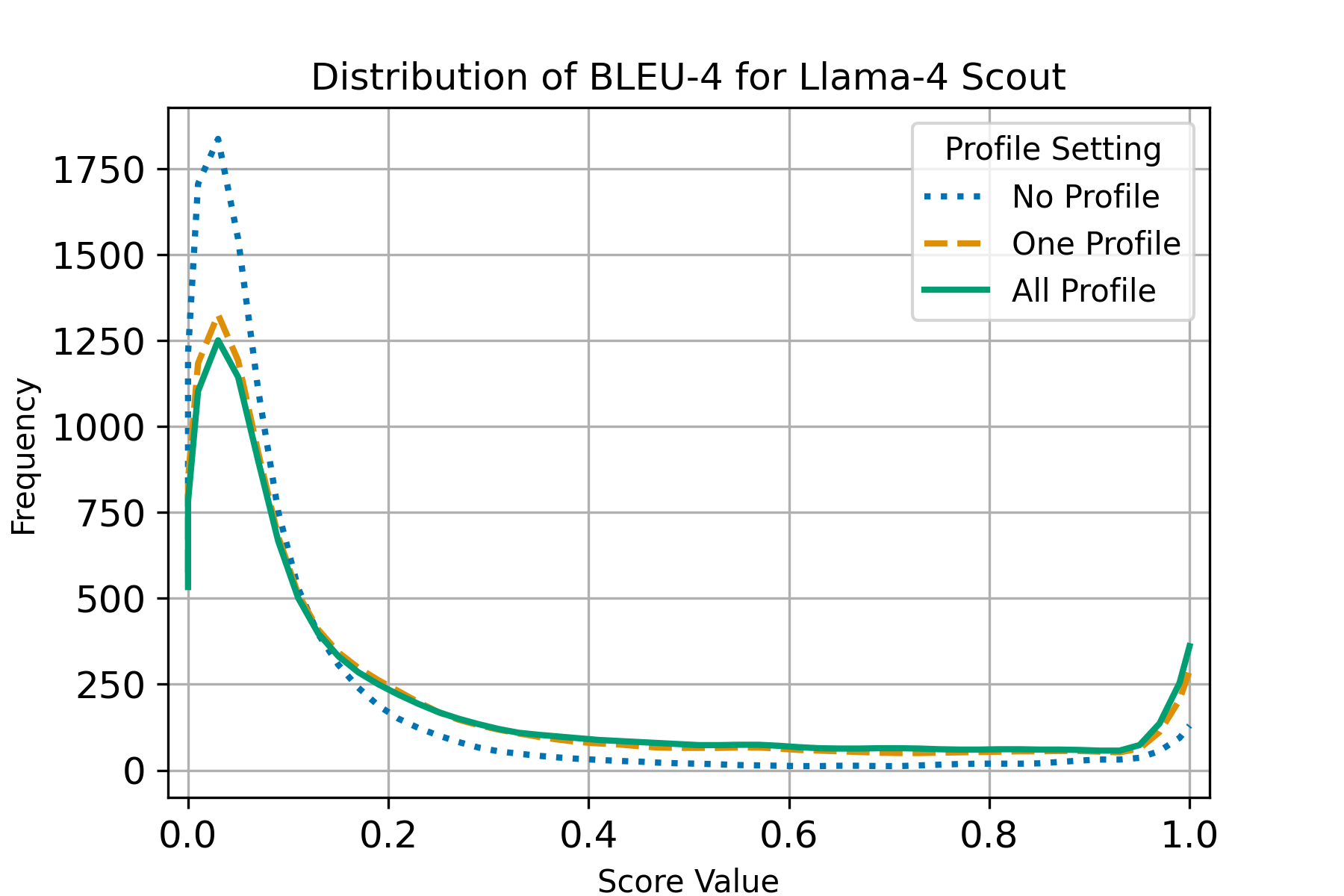}
        \vspace{-1.5pc}
        \caption{BLEU-4 for Llama-4 Scout}
        \label{fig:top_right-bleu4}
    \end{subfigure}
    
    \begin{subfigure}{0.495\textwidth}
        \includegraphics[width=\textwidth]{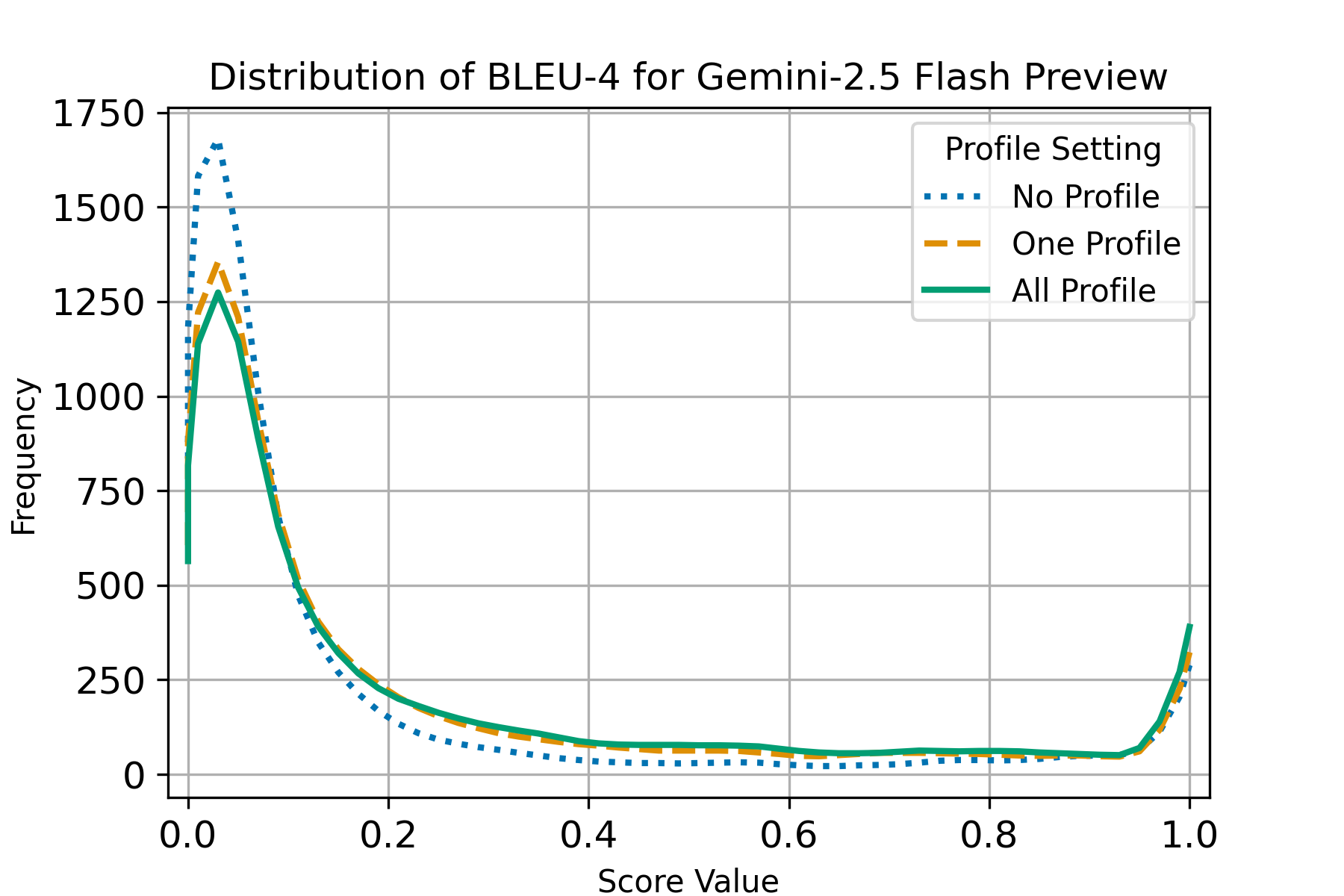}
        \vspace{-1.5pc}
        \caption{BLEU-4 for Gemini-2.5 Flash Preview}
        \label{fig:bottom_left-bleu4}
    \end{subfigure}
    \hfill
    \begin{subfigure}{0.495\textwidth}
        \includegraphics[width=\textwidth]{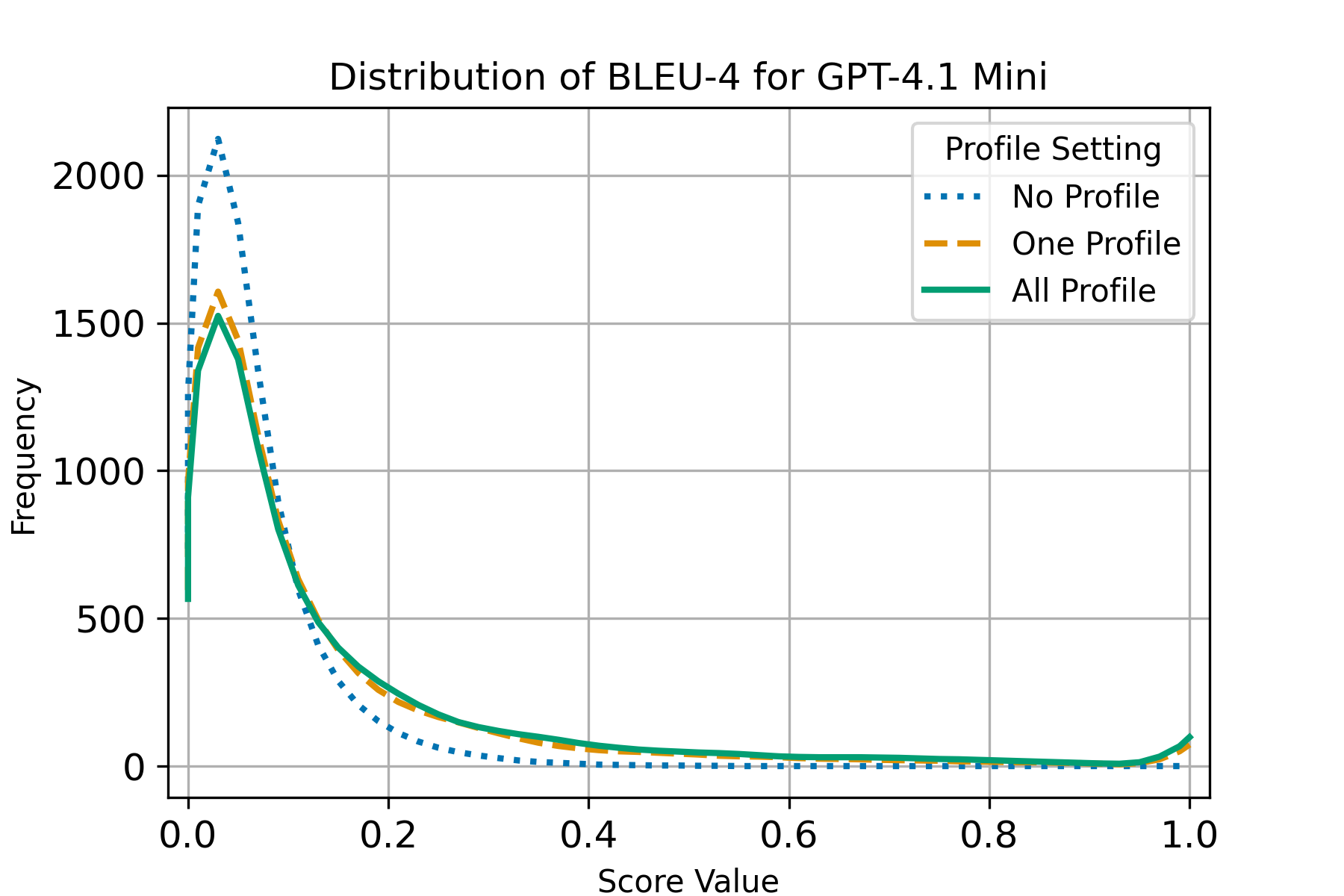}
        \vspace{-1.5pc}
        \caption{BLEU-4 for GPT-4.1 Mini}
        \label{fig:bottom_right-bleu4}
    \end{subfigure}
    \caption{Distribution of the BLEU-4 across different LLMs and profile configuration.}
    \label{fig:dist-bleu4}
\end{figure*}

\begin{figure*}[t]
    \centering
    \begin{subfigure}{0.495\textwidth}
        \includegraphics[width=\textwidth]{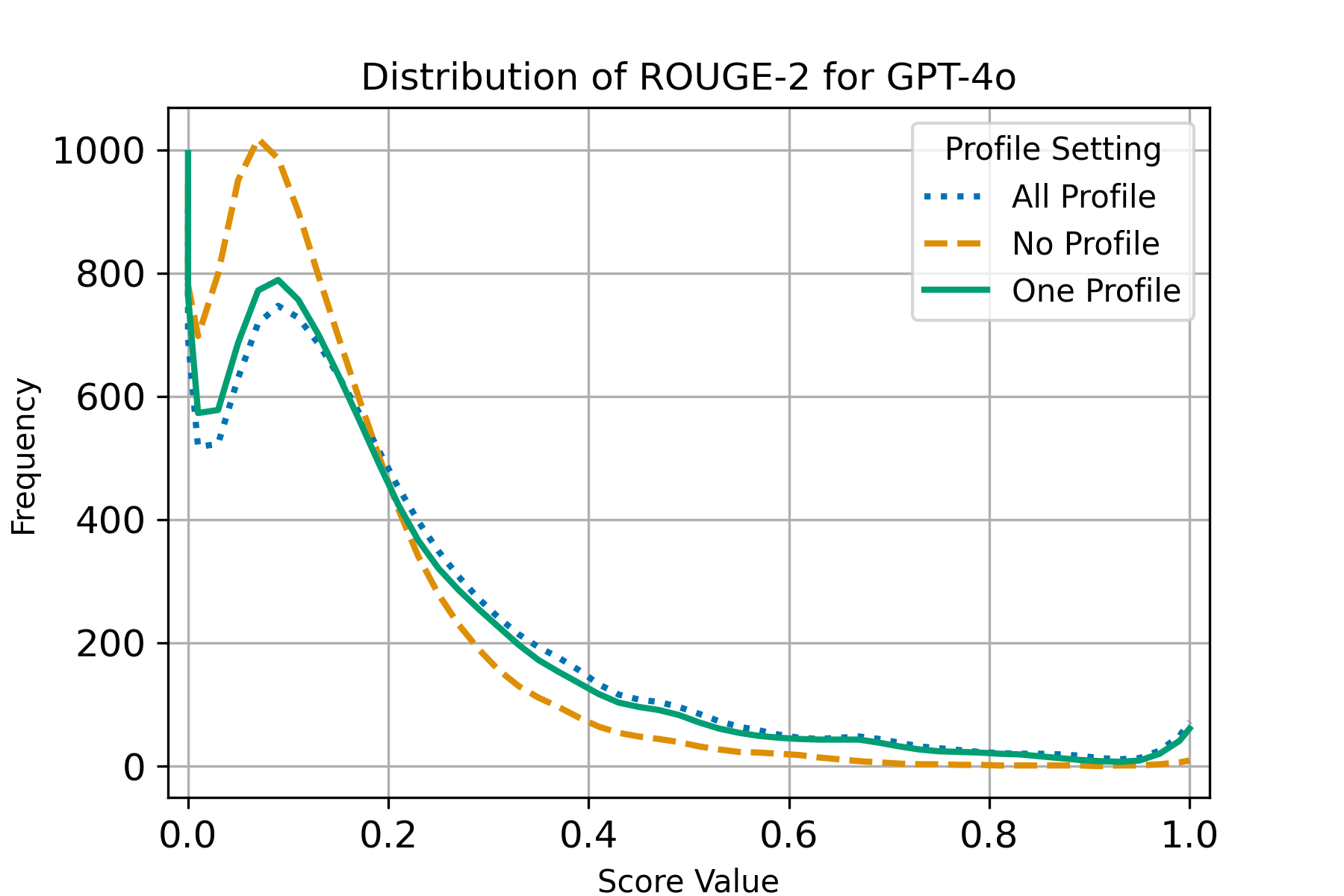}
        \vspace{-1.5pc}
        \caption{ROUGE-2 for GPT-4o}
        \label{fig:top_left-rouge2}
    \end{subfigure}
    \hfill
    \begin{subfigure}{0.495\textwidth}
        \includegraphics[width=\textwidth]{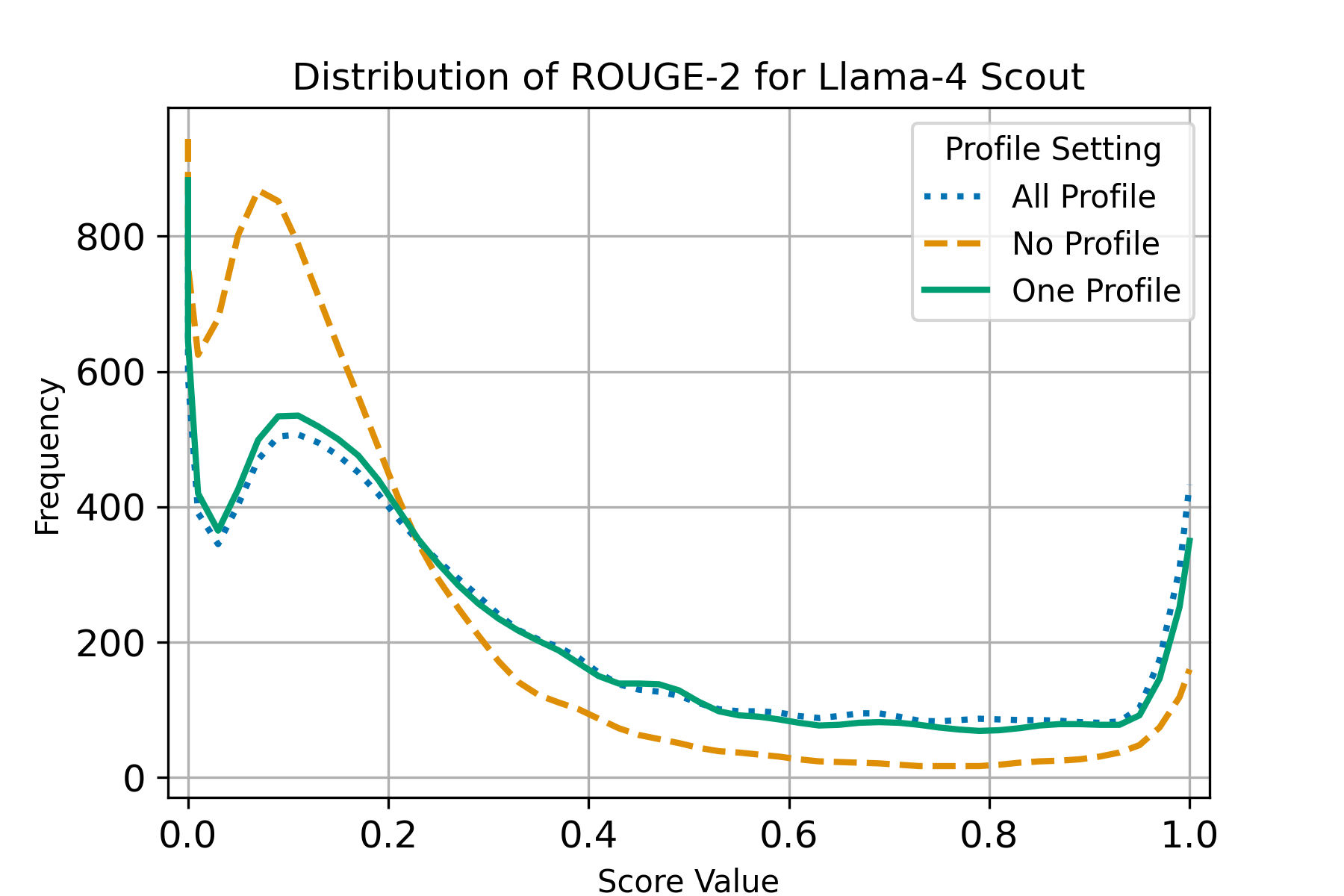}
        \vspace{-1.5pc}
        \caption{ROUGE-2 for Llama-4 Scout}
        \label{fig:top_right-rouge2}
    \end{subfigure}
    
    \begin{subfigure}{0.495\textwidth}
        \includegraphics[width=\textwidth]{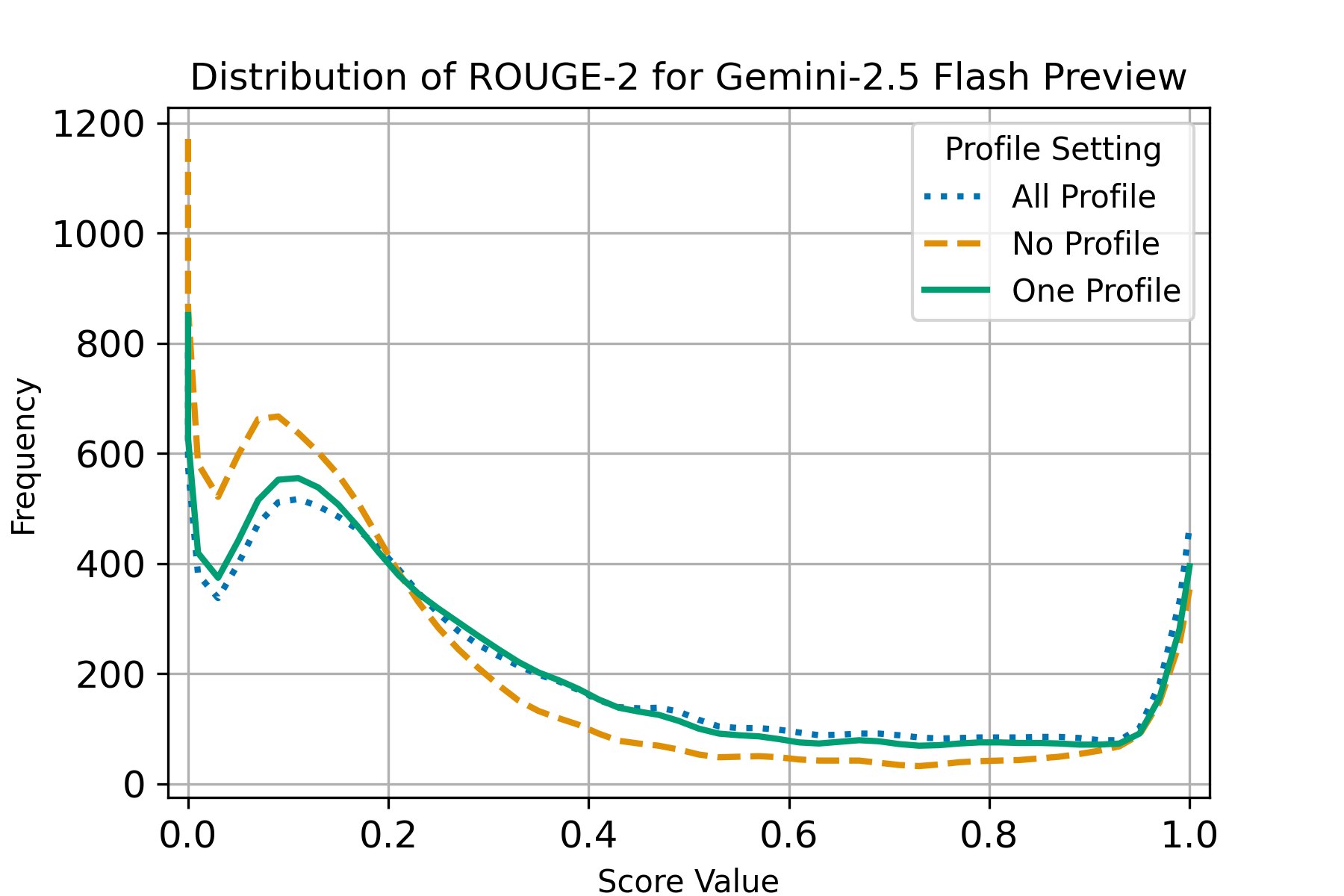}
        \vspace{-1.5pc}
        \caption{ROUGE-2 for Gemini-2.5 Flash Preview}
        \label{fig:bottom_left-rouge2}
    \end{subfigure}
    \hfill
    \begin{subfigure}{0.495\textwidth}
        \includegraphics[width=\textwidth]{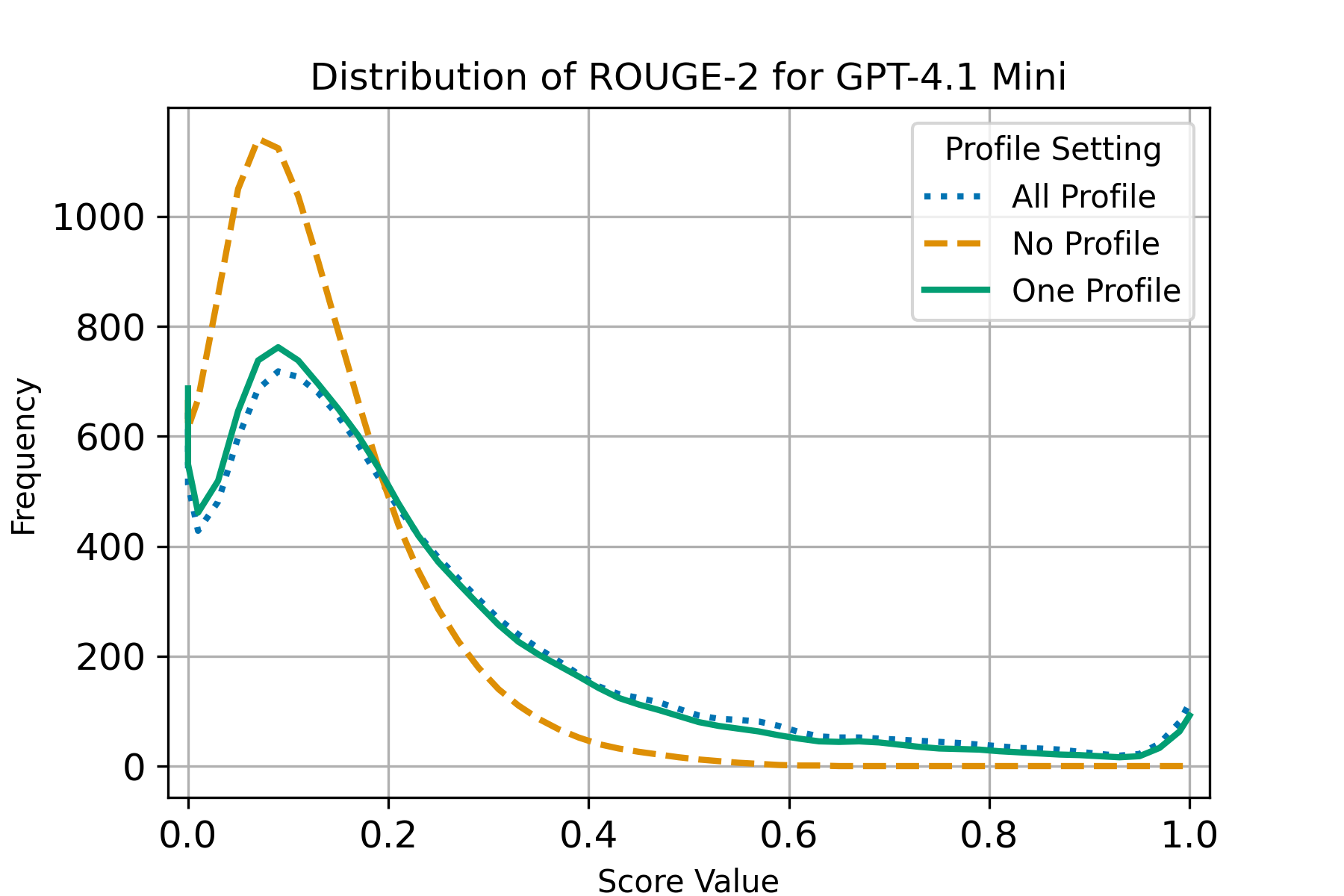}
        \vspace{-1.5pc}
        \caption{ROUGE-2 for GPT-4.1 Mini}
        \label{fig:bottom_right-rouge2}
    \end{subfigure}
    \caption{Distribution of the ROUGE-2 across different LLMs and profile configuration.}
    \label{fig:dist-rouge2}
\end{figure*}

This appendix session provides supplementary materials for the human evaluation study described in \autoref{sec:humanEval}. \autoref{fig:human-eval} shows the interface of the user study in Microsoft Form. 

To supplement the 

Our protocol was reviewed and approved by the Institutional Review Board (IRB) of The Pennsylvania State University (STUDY00025214), which granted a waiver of written documentation of consent. Consent was implied by participants voluntary action of proceeding with and completing the survey.
Each participant received \$20 cash compensation upon completion of the study. The following is the instruction shown to the human participant before they start the study.

\begin{lstlisting}[basicstyle=\small\ttfamily, breaklines=true, frame=single]
We are conducting a human evaluation study on the quality of captions generated for scientific figures.
For each question, you will be shown the paper's title and abstract to provide context, followed by a figure and four caption options.
Your task is to rank the captions based on how well they help you understand the figure.

Some captions may be generated with the assistance of AI. However, the goal is not to identify which captions are human- or AI-written. Please focus only on the clarity and usefulness of each caption in conveying the figure's message.

There are no right or wrong answers; please use your own judgment. The survey includes 50 figures and takes about 1.5 hours to complete. Please use a desktop computer only. To ensure fair evaluation, please avoid searching for the original papers online while completing the task.
\end{lstlisting}

And for each figure, we ask them to do the ranking of the captions with the following prompt:
\begin{lstlisting}[basicstyle=\small\ttfamily, breaklines=true, frame=single]
Please rank the four captions below based on how well they help you understand the figure.
Drag and drop to reorder them from 1 (best) to 4 (worst) using your mouse.
\end{lstlisting}

\section{Disclosure of AI Assistance\label{app:ai-assistance}}
We used Perplexity and Gemini to facilitate proofreading and text refinement.

\begin{table}[t]
\centering
\small
\renewcommand{\arraystretch}{1.2}
\setlength{\tabcolsep}{1pt}
\definecolor{lightgreen}{rgb}{0.9, 1.0, 0.9}
\definecolor{darkgreen}{rgb}{0.7, 0.9, 0.7}

\begin{tabular}{@{}llll@{}}
\toprule
\textbf{Model}                                                      & \textbf{No-Profile} & \cellcolor{lightgreen}\textbf{1-Profile} & \cellcolor{darkgreen}\textbf{All-Profile} \\ \midrule
GPT-4o                                                              & .844                & \cellcolor{lightgreen}.860               & \cellcolor{darkgreen}.863                 \\ \midrule
Llama-4 Scout                                                       & .856                & \cellcolor{lightgreen}.873               & \cellcolor{darkgreen}.876                 \\ \midrule
\begin{tabular}[c]{@{}l@{}}Gemini-2.5 \\ Flash Preview\end{tabular} & .865                & \cellcolor{lightgreen}.874               & \cellcolor{darkgreen}.877                 \\ \midrule
GPT-4.1 Mini                                                        & .844                & \cellcolor{lightgreen}.860               & \cellcolor{darkgreen}.863                 \\ \bottomrule
\end{tabular}

\caption{Performance of various LLMs on figure caption generation, as measured by BERTScore.}
\label{tab:BERTScore_LLM}
\end{table}

\end{document}